\documentclass{article}
\usepackage{ijcai23}

\usepackage{booktabs}
\usepackage[export]{adjustbox}
\usepackage{multirow}
\usepackage{multicol}
\usepackage[table,xcdraw]{xcolor}
\usepackage{times}
\usepackage{soul}
\usepackage{url}
\usepackage[hidelinks]{hyperref}
\usepackage[utf8]{inputenc}
\usepackage[small]{caption}
\usepackage{graphicx}
\usepackage{amsmath}
\usepackage{amsthm}
\usepackage{booktabs}
\usepackage{algorithm}
\usepackage{algorithmic}

\usepackage[switch]{lineno}

\title{A Large-scale Film Style Dataset for Learning Multi-frequency Driven Film Enhancement}

\author{
Zinuo Li$^1$\thanks{Equal Contribution}
\and
Xuhang Chen$^1$$^2$\footnotemark[1] \and
Shuqiang Wang$^2$\thanks{Corresponding author}\and
Chi-Man Pun$^1$\footnotemark[2]
\affiliations
$^1$University of Macau\\
$^2$Shenzhen Institute of Advanced Technology, Chinese Academy of Sciences
\emails
cmpun@umac.mo
}

\begin{document}

\maketitle

\begin{abstract}
Film, a classic image style, is culturally significant to the whole photographic industry since it marks the birth of photography.
However, film photography is time-consuming and expensive, necessitating a more efficient method for collecting film-style photographs. 
Numerous datasets that have emerged in the field of image enhancement so far are not film-specific.
In order to facilitate film-based image stylization research, we construct FilmSet, a large-scale and high-quality film style dataset. Our dataset includes three different film types and more than 5000 in-the-wild high resolution images. 
Inspired by the features of FilmSet images, we propose a novel framework called FilmNet based on Laplacian Pyramid for stylizing images across frequency bands and achieving film style outcomes. Experiments reveal that the performance of our model is superior than state-of-the-art techniques. The link of code and data is \url{https://github.com/CXH-Research/FilmNet}.
\end{abstract}

\section{Introduction}

Film imaging is a special chemical process~\cite{teubner2019optical} that generates a unique color and graininess, which is different from that of digital cameras. It is the graininess of film-style images that gives the whole picture a unique charm. The beauty shown in film photographs has demonstrated its charm. Hence, film has become a prominent kind of photography in the minds of many. Years of adjusting by film makers have made it possible to present films in colors that meet the aesthetics of the public. As a result, people have a better sense of the colors presented by film, which has significant implications for the field of enhancing the beauty of images. 

\begin{figure}[!ht]
    \begin{minipage}[b]{1.0\linewidth}
        \begin{minipage}[b]{.49\linewidth}
            \centering
            \centerline{\includegraphics[width=\linewidth]{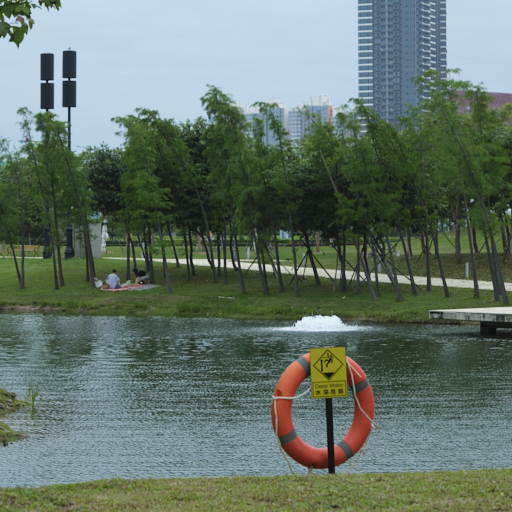}}
            \centerline{(a) Input}\medskip
        \end{minipage}
        \hfill
        \begin{minipage}[b]{0.49\linewidth}
            \centering
            \centerline{\includegraphics[width=\linewidth]{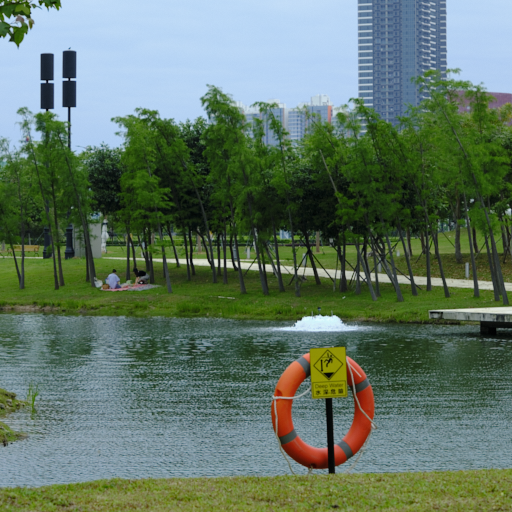}}
            \centerline{(b) Film style}\medskip
        \end{minipage}
    \end{minipage}
    
    \begin{minipage}[b]{1.0\linewidth}
        \begin{minipage}[b]{0.49\linewidth}
            \centering
            \centerline{\includegraphics[width=\linewidth]{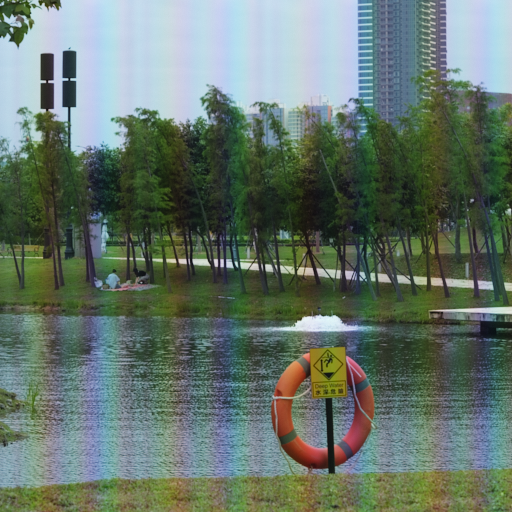}}
            \centerline{(c) Result of STAR-DCE}\medskip
        \end{minipage}
        \hfill
        \begin{minipage}[b]{0.49\linewidth}
            \centering
            \centerline{\includegraphics[width=\linewidth]{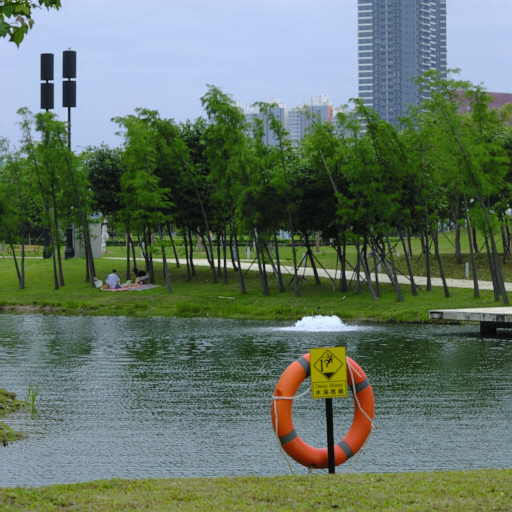}}
            \centerline{(d) Result of Ours}\medskip
        \end{minipage}
    \end{minipage}
    \caption{
    This figure contains input image~(a) and film style image~(b) from our FilmSet. Although existing deep learning image enhancement methods such as STAR-DCE~(c) may not perform well, our method~(d) can properly enhance the image towards film style.
    }
    \label{fig:intro}
\end{figure}

Although film style is appealing, film photography is time-consuming, labor-intensive, and expensive. Therefore, many individuals begin to minimize the professionalism of film photography to save time and funds by digitally simulating film styles. Designed to replace tedious human labor, the Look-Up-Table~(LUT) is a reliable tool for automatic image color grading. The fundamental premise underlying them is transforming input into a certain output value using efficient lookup and interpolation algorithms. In recent years, deep learning has pushed the development of LUTs, resulting in an explosion of fascinating research~\cite{wang2022lcdp,liu2022degradation,wang2022neural,song2021starenhancer,kim2021representative,kim2020global,yang2022seplut}.

Despite the above-mentioned studies, film stylization has not yet been investigated. A great number of individuals have been drawn to film's timeless and alluring visual qualities. Nevertheless, many old film cameras lack the ability to export digital images, and digital cameras cannot replicate film imagery so that they must rely on LUT for film simulation. 

\begin{figure*}[!ht]
    \begin{minipage}[b]{1.0\linewidth}
        \begin{minipage}[b]{0.5\linewidth}
            \centering
            \centerline{\includegraphics[width=\linewidth]{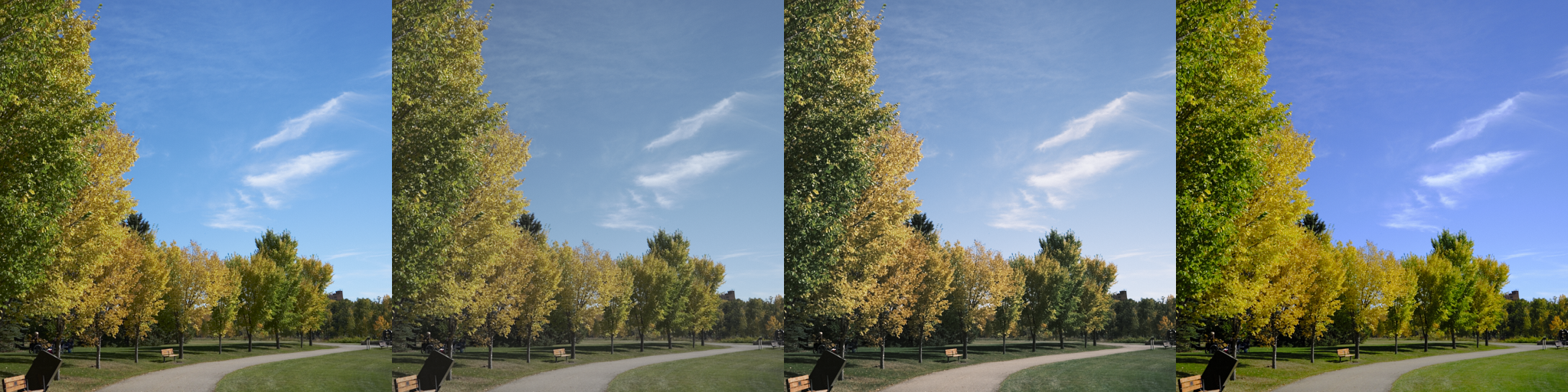}}
        \end{minipage}
        \begin{minipage}[b]{0.5\linewidth}
            \centering
            \centerline{\includegraphics[width=\linewidth]{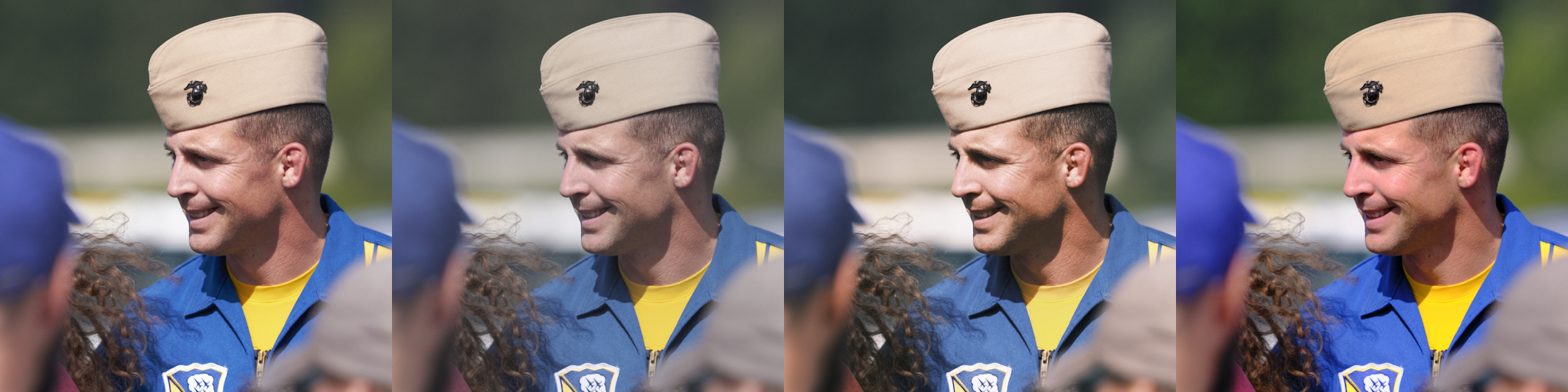}}
        \end{minipage}
    \end{minipage}
    \begin{minipage}[b]{1.0\linewidth}
        \begin{minipage}[b]{0.5\linewidth}
            \centering
            \centerline{\includegraphics[width=\linewidth]{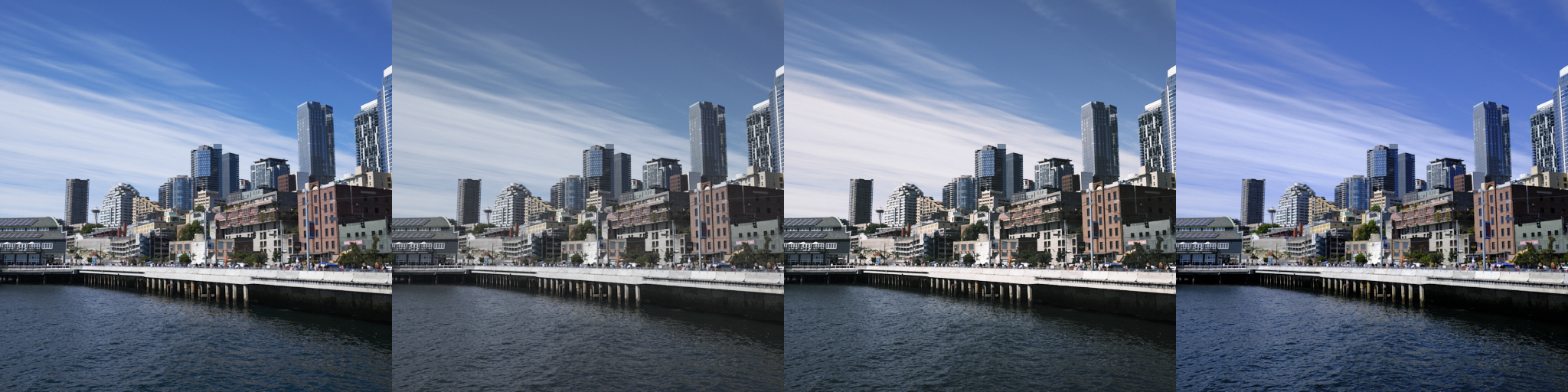}}
        \end{minipage}
        \begin{minipage}[b]{0.5\linewidth}
            \centering
            \centerline{\includegraphics[width=\linewidth]{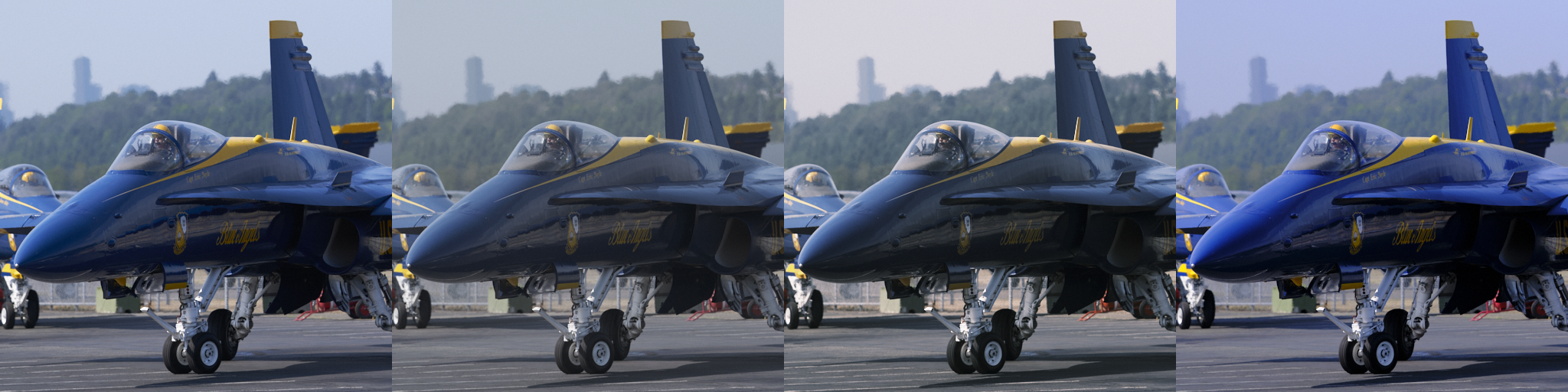}}
        \end{minipage}
    \end{minipage}

    
    \caption{
    Visual samples demonstrating the variety of the proposed dataset FilmSet, such as various scenes, portraits and film types. Each image depicts the original image, the Cinema style, the Classic Negative style, and the Velvia style horizontally.
    }
    \label{fig:dataset}
\end{figure*}

The challenge is that the current LUTs or some methods focus mostly on global color and lighting, while other detailed operations may get less consideration. Consequently, the visual quality may deteriorate and they may become invalid, as shown in Figure~\ref{fig:intro}~(c). In addition, the existing image enhancement datasets MIT FiveK~\cite{bychkovsky2011learning} and HDR Plus~\cite{hasinoff2016burst} are designed for broad use and do not adequately address our issue. 

Therefore, our objective is to extract additional features and enhance photographs to resemble film. Thus, we construct a large scale film-specific dataset that allows us to facilitate relative film style research, namely the FilmSet. We found that the features of film style images are very suitable for the enhancement in multi-frequency, inspired this, we propose a novel framework utilizing Laplacian Pyramid~\cite{burt1987laplacian}. We summarize our major contributions as follows: 

\begin{enumerate}
    \item We are the first to construct a large-scale high-quality dataset with 3 groups of different film style and a total of 5,285 high-quality images, called FilmSet.
    \item To learn the features in FilmSet properly, we present FilmNet, a novel multi-frequency framework based on Laplacian Pyramid for simulating film styles and subsequently retouching normal photos.
    \item We demonstrate our model is superior to the state-of-the-art methods via extensive experiments on our dataset and other publicly accessible benchmark datasets.
\end{enumerate}

\section{Related Work}

\subsection{Lookup tables}

A LUT is an array that supplants run-time calculation with a more straightforward array indexing process. Once the LUT is created, input images can be retouched using only the memory access and interpolation without further recalculation. 

Previous works focus on mastering LUTs to simulate the color adjustment curves of well-known picture editing software~\cite{song2021starenhancer,kim2021representative,guo2020zero,bianco2020personalized}. By learning a large number of image-independent basic LUTs and combining them with image-dependent weights, it is possible to predict LUTs that are adaptable to a variety of picture contents. Therefore, it is viable to use digital LUT to replicate the film imaging process. However, more refinements are needed to learn the features properly instead of simply using a single LUT, such as refining in different frequency bands or
focus both on detailed and global features.

\subsection{Photo retouching methods}
Previously, experts and professional image editing systems are required for photo retouching to optimize global tuning and adjust local aspects. Nowadays, deep learning models are widely used to retouch photos. 

Inspired by bilateral grid processing and local affine color transforms, HDRNet is proposed to use in smartphones~\cite{gharbi2017deep}. Then, Hui Zeng \emph{et al.} proposed to learn 3D LUTs from annotated data using pairwise or unpaired learning~\cite{zeng2020learning}. SepLUT~\cite{yang2022seplut}, separated a single color transform into component-independent and component-correlated sub-transforms to enhance images. Liang \emph{et al.} propose LPTN based on Laplacian Pyramid~\cite{liang2021high}. Nonetheless, few studies have concentrated on film stylization. 

\subsection{Image enhancement datasets}

Recently, there has appeared many datasets that enable learning photo enhancement and retouching. MIT FiveK~\cite{bychkovsky2011learning} is a general-purpose dataset which consists of 5,000 original images of broad situations and five versions of retouched targets. Another example is HDR Plus~\cite{hasinoff2016burst}. HDR Plus is a burst photography dataset which consists of 3640 bursts (made up of 28461 images in total), organized into sub-folders.

Despite these significant efforts, the preceding datasets were constructed in general scenarios and film-style photographs were not included. Therefore, the models trained on them are inappropriate for the film stylization task. In this study, we produce an expansive FilmSet dataset to help this endeavor.

\section{FilmSet Dataset}

As previously stated, current datasets and models for photo retouching cannot meet the needs of film stylization. To address these issues, we develop a vast and high-quality dataset, called FilmSet. Visual samples are available in Figure~\ref{fig:dataset}. We have three syles in total: Cinema, Classical Negative (ClassNeg) and Velvia, each style contains 5285 images.

\subsection{Challenges}

To develop a worthwhile film-style dataset that meets real-world needs, we must surmount a number of obstacles. First, the photographs should be in high-quality raw format. In contrast to the ubiquitous and widely accessible JPG photographs, raw photos are far more difficult to get. Second, the dataset should be extensive and encompass a broad variety of real-world scenarios in terms of shooting purpose, diverse settings and portraits, lighting circumstances, and film type, hence increasing the cost of data collecting.

\begin{figure*}[ht]
    \begin{minipage}[b]{1.0\linewidth}
        \includegraphics[width=\linewidth]{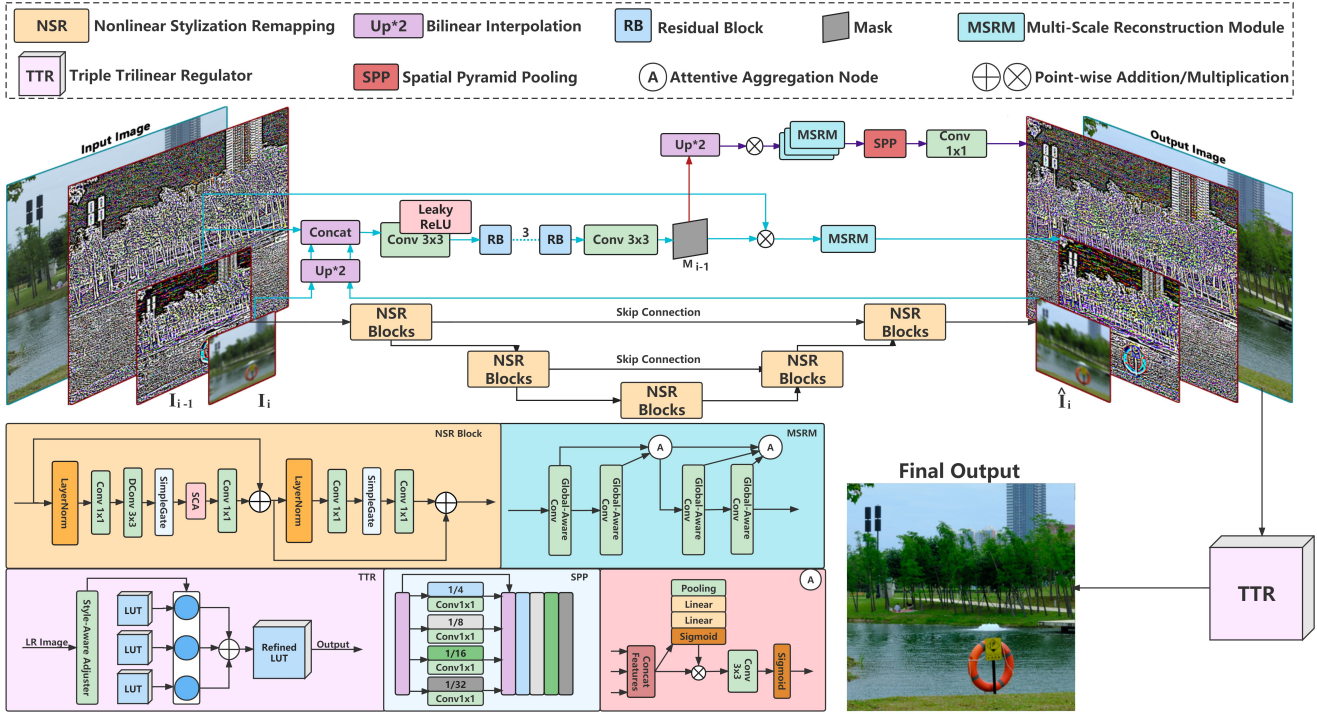}
    \end{minipage}
    \caption{
    The general structure of our FilmNet network. Initially, the LP divides the input picture into three frequency bands and sends them to distinct networks. These pieces are combined into the output picture before being transmitted to TTR. Eventually, the TTR performs the final output processing.
    }
    \label{fig:filmnet}
\end{figure*}

\subsection{Data collection and selection}

To gather as many raw film-style photographs as feasible, we collected license-free samples from individual photographers and professional photography studios. Additionally, we supplemented some raw format images taken by ourselves.

When gathering data, we carefully inspected the variety of raw images in terms of the shooting occasion, the portrait, background scenes and other possible variants. Figure~\ref{fig:dataset} illustrates the variety of images obtained.

\subsection{Film simulation}

Initially, we amassed over 8000 raw images, after which we undertook multiple rounds of curation. We initially discarded photos with poor quality, such as significant motion blur or out-of-focus, as well as those containing improper information. In addition, we meticulously examined photos group by group, eliminating outliers and duplicates. We acquired a total of 5285 photographs after the screening. 

Using Capture One~\cite{c1}, we independently applied three film recipes to each of these 5000 images as ground truth. Fujifilm is renowned for its world-class film manufacturing and quality, and Capture One's film stylization LUT imitates three Fujifilm film styles flawlessly due to their extensive collaboration.

\begin{figure*}[!ht]
    \begin{minipage}[b]{1.0\linewidth}
        \begin{minipage}[b]{0.12\linewidth}
            \centering
            \centerline{\includegraphics[height=4.2cm]{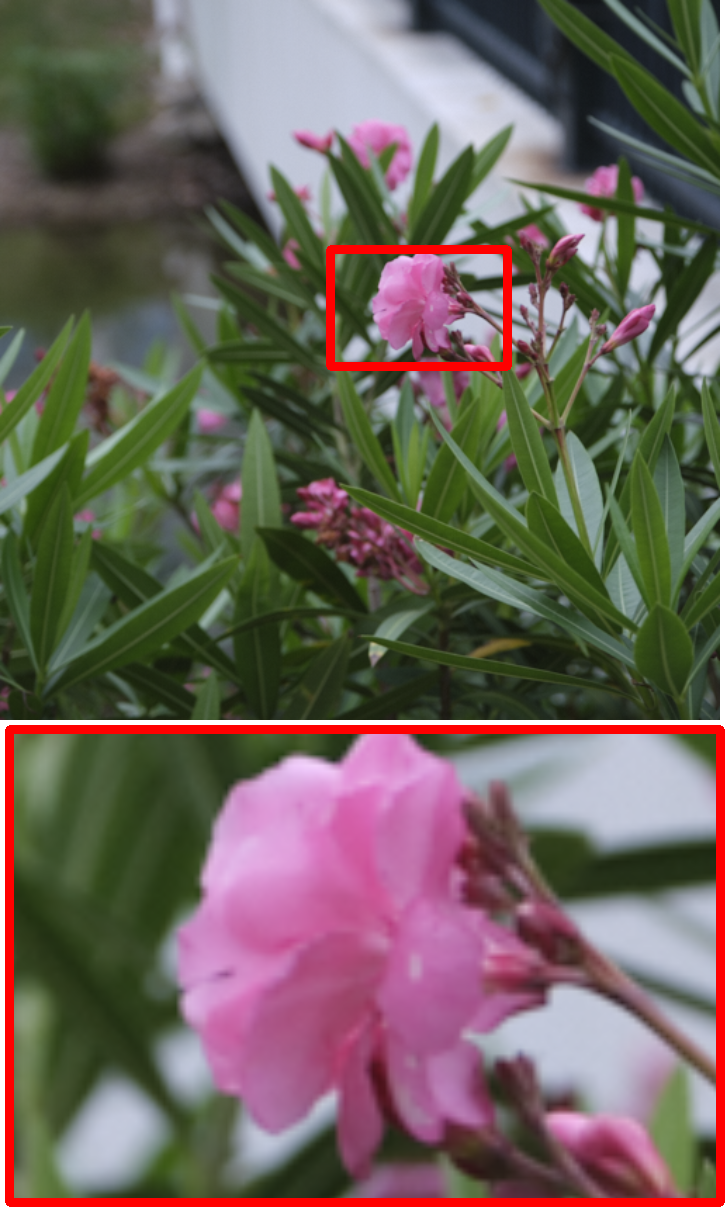}}
        \end{minipage}
        \hfill
        \begin{minipage}[b]{0.12\linewidth}
            \centering
            \centerline{\includegraphics[height=4.2cm]{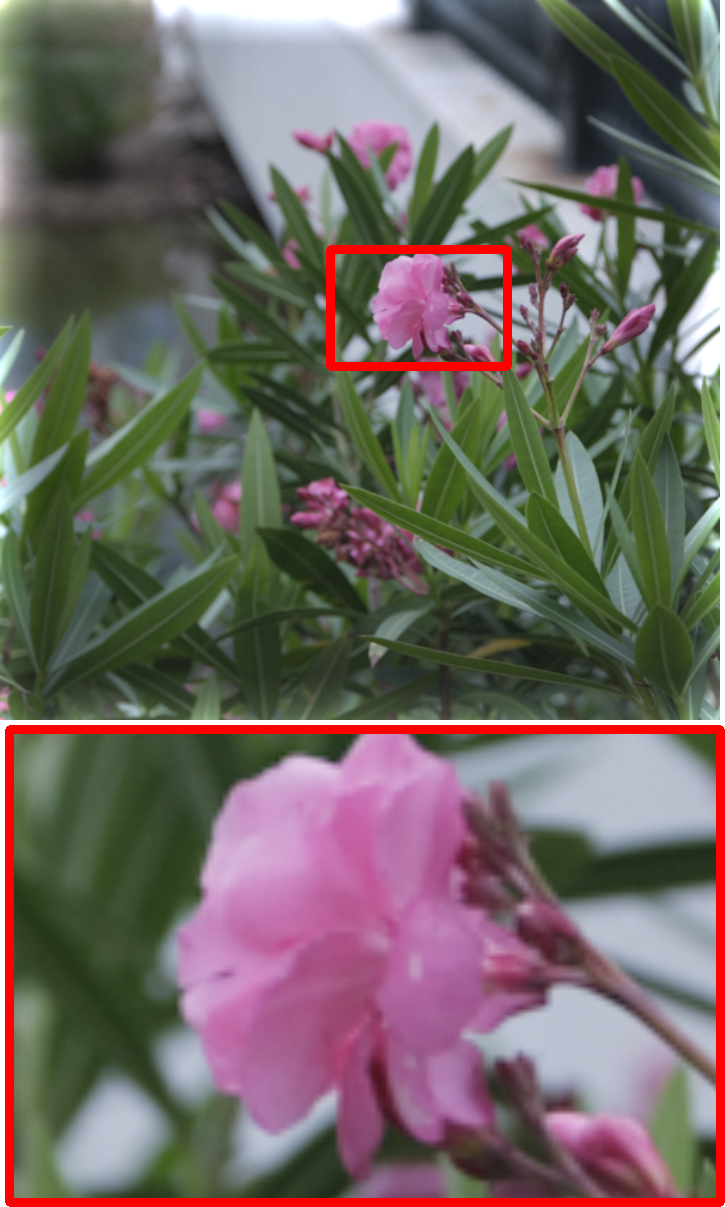}}
        \end{minipage}
        \hfill
        \begin{minipage}[b]{0.12\linewidth}
            \centering
            \centerline{\includegraphics[height=4.2cm]{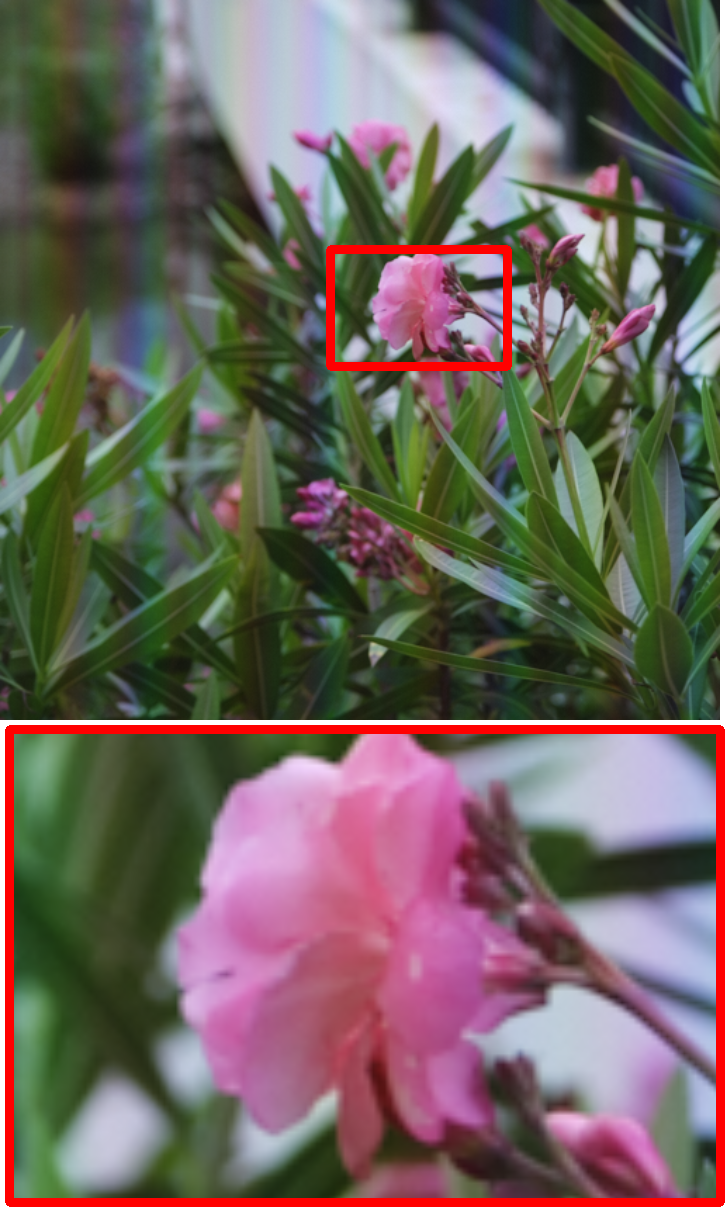}}
        \end{minipage}
        \hfill
        \begin{minipage}[b]{.12\linewidth}
            \centering
            \centerline{\includegraphics[height=4.2cm]{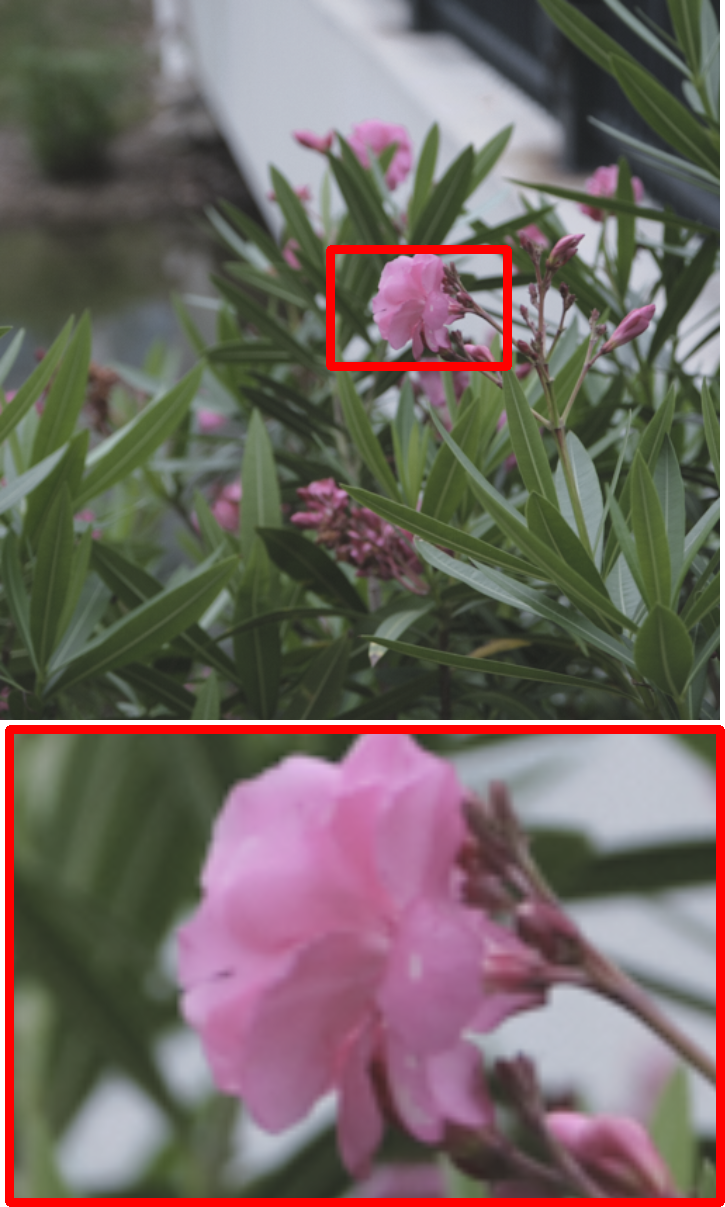}}
        \end{minipage}
        \hfill
        \begin{minipage}[b]{0.12\linewidth}
            \centering
            \centerline{\includegraphics[height=4.2cm]{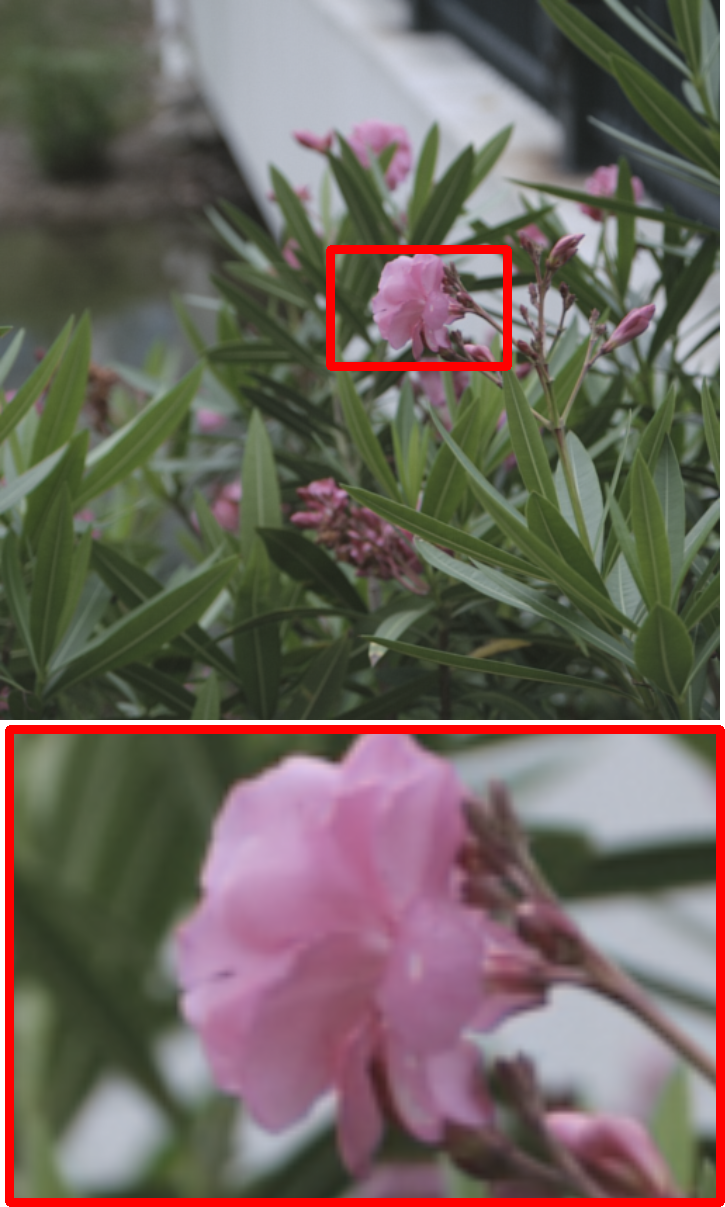}}
        \end{minipage}
        \hfill
        \begin{minipage}[b]{0.12\linewidth}
            \centering
            \centerline{\includegraphics[height=4.2cm]{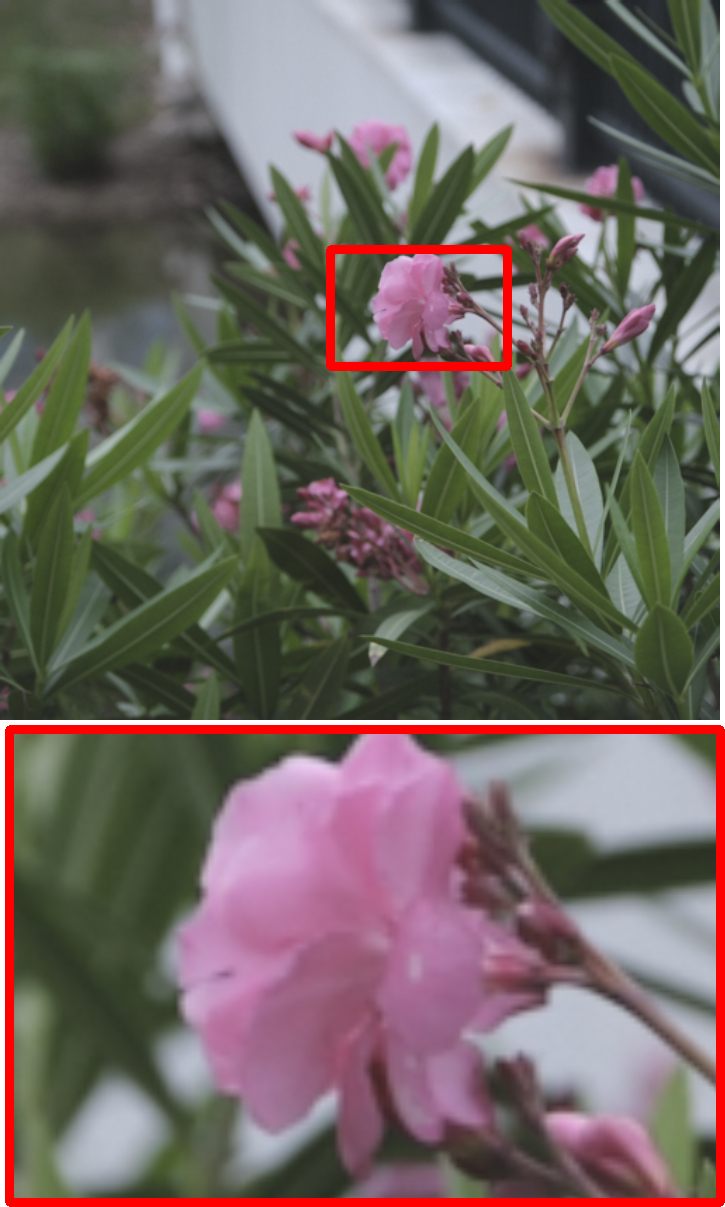}}
        \end{minipage}
        \hfill
        \begin{minipage}[b]{0.12\linewidth}
            \centering
            \centerline{\includegraphics[height=4.2cm]{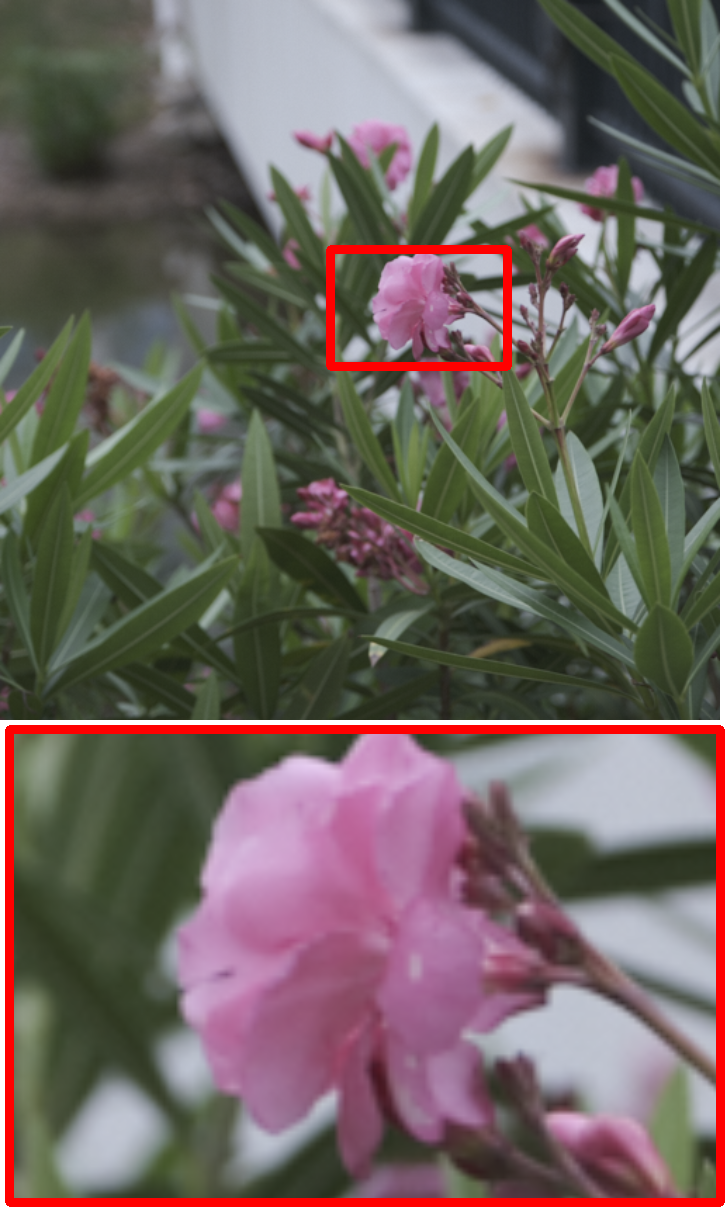}}
        \end{minipage}
    \end{minipage}

    \begin{minipage}[b]{1.0\linewidth}
        \begin{minipage}[b]{0.12\linewidth}
            \centering
            \centerline{\includegraphics[height=4.2cm]{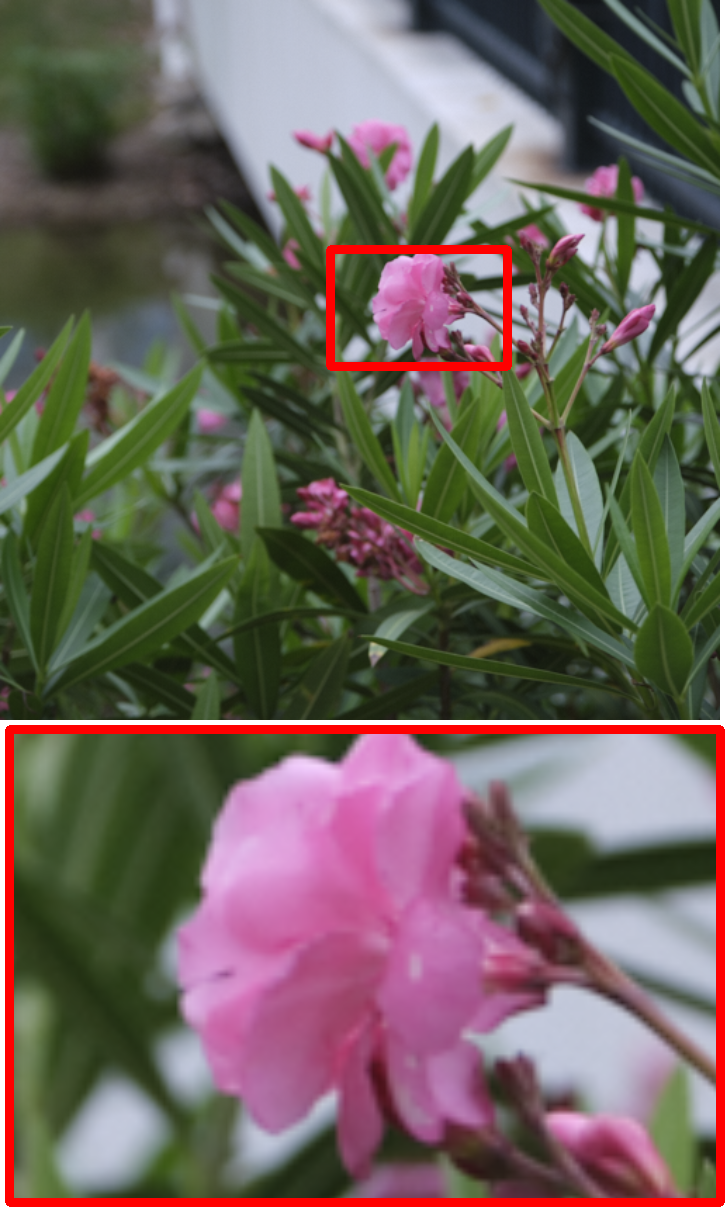}}
        \end{minipage}
        \hfill
        \begin{minipage}[b]{0.12\linewidth}
            \centering
            \centerline{\includegraphics[height=4.2cm]{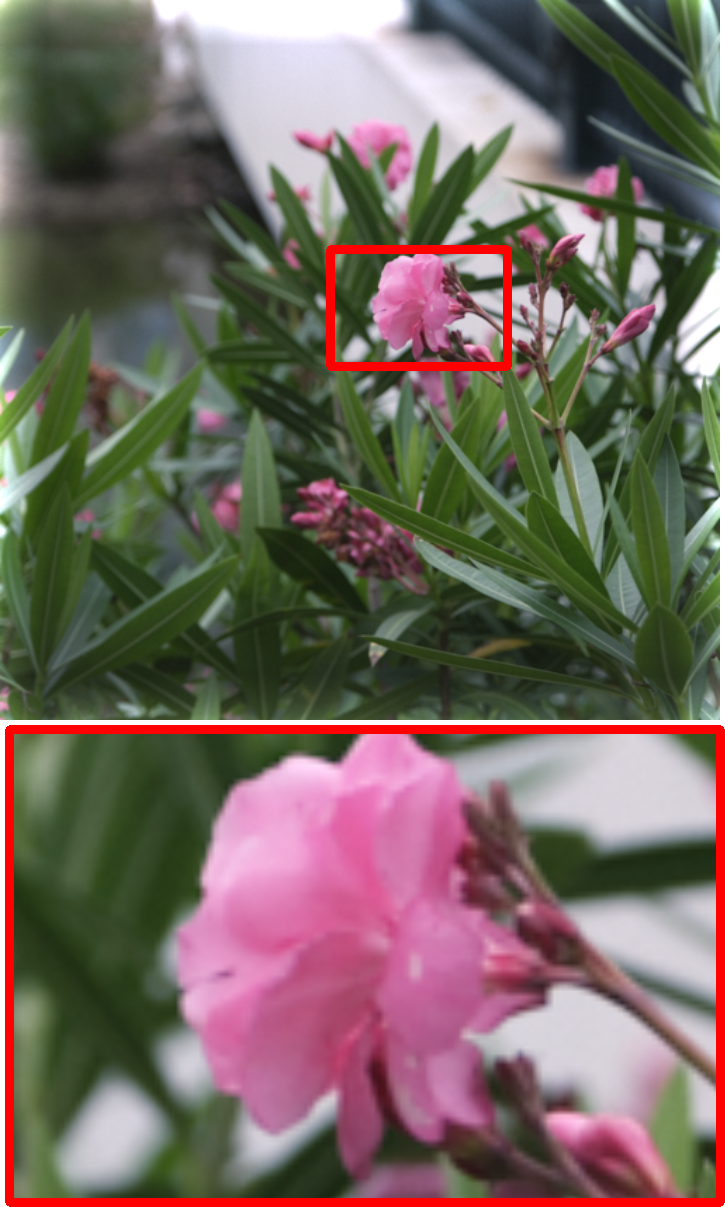}}
        \end{minipage}
        \hfill
        \begin{minipage}[b]{0.12\linewidth}
            \centering
            \centerline{\includegraphics[height=4.2cm]{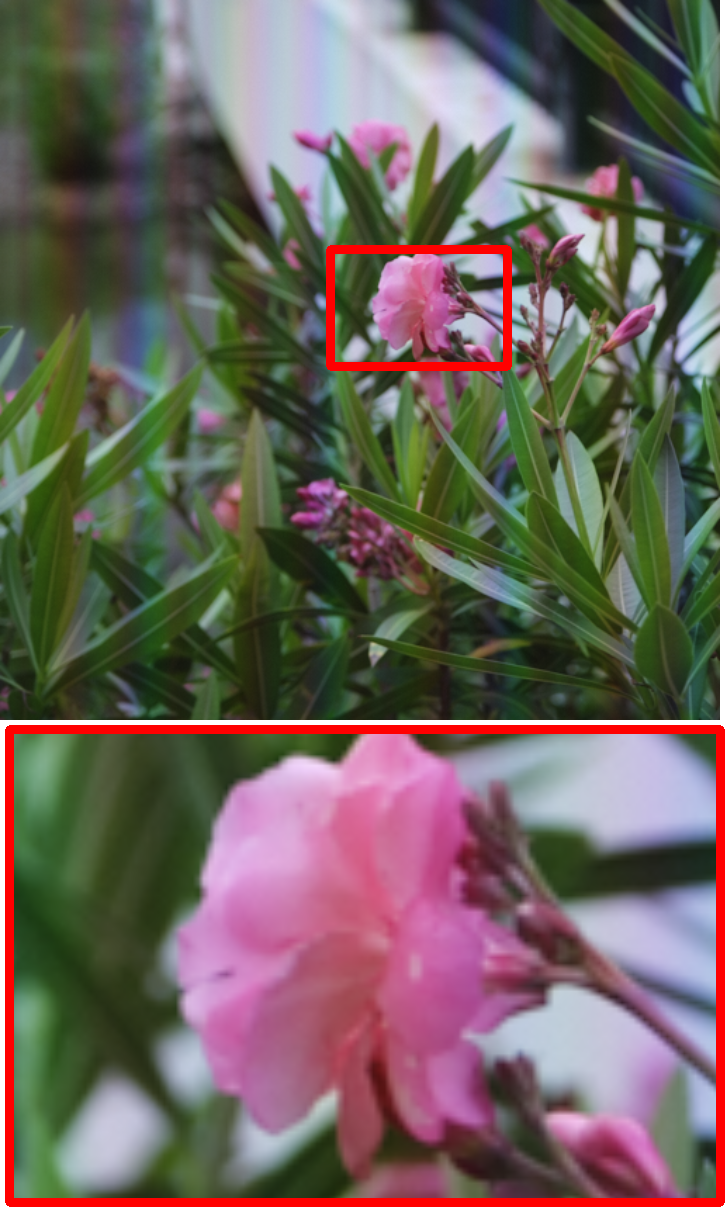}}
        \end{minipage}
        \hfill
        \begin{minipage}[b]{.12\linewidth}
            \centering
            \centerline{\includegraphics[height=4.2cm]{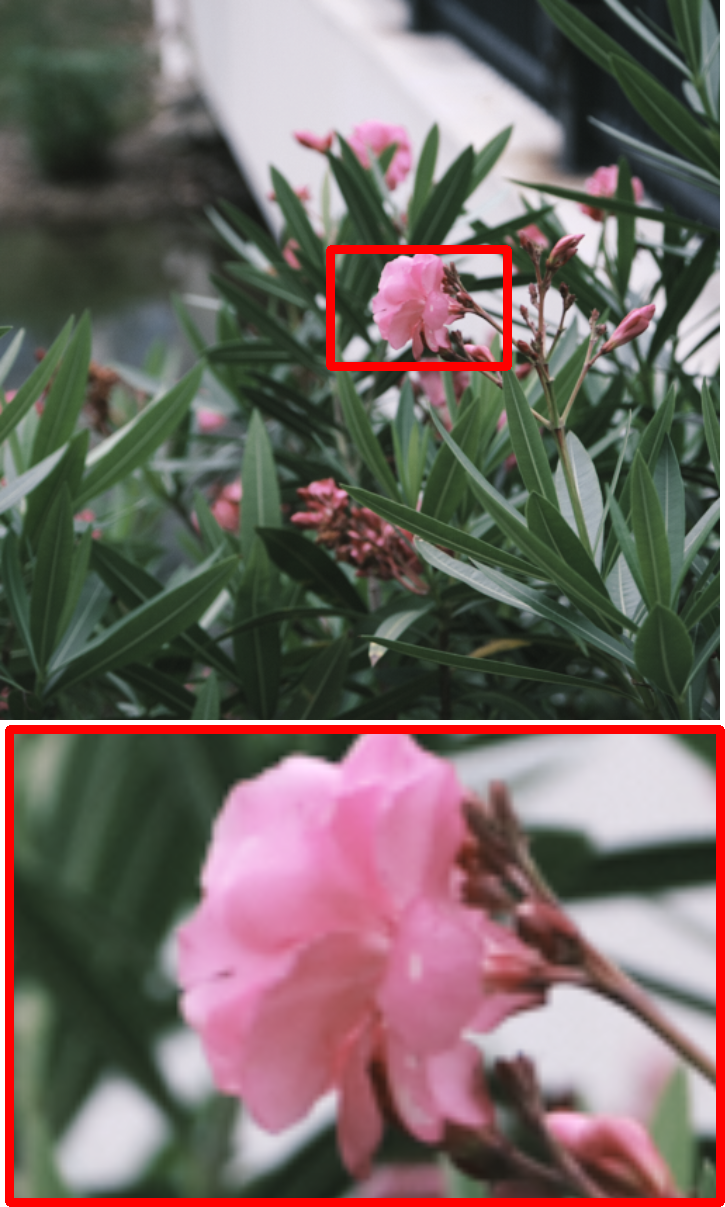}}
        \end{minipage}
        \hfill
        \begin{minipage}[b]{0.12\linewidth}
            \centering
            \centerline{\includegraphics[height=4.2cm]{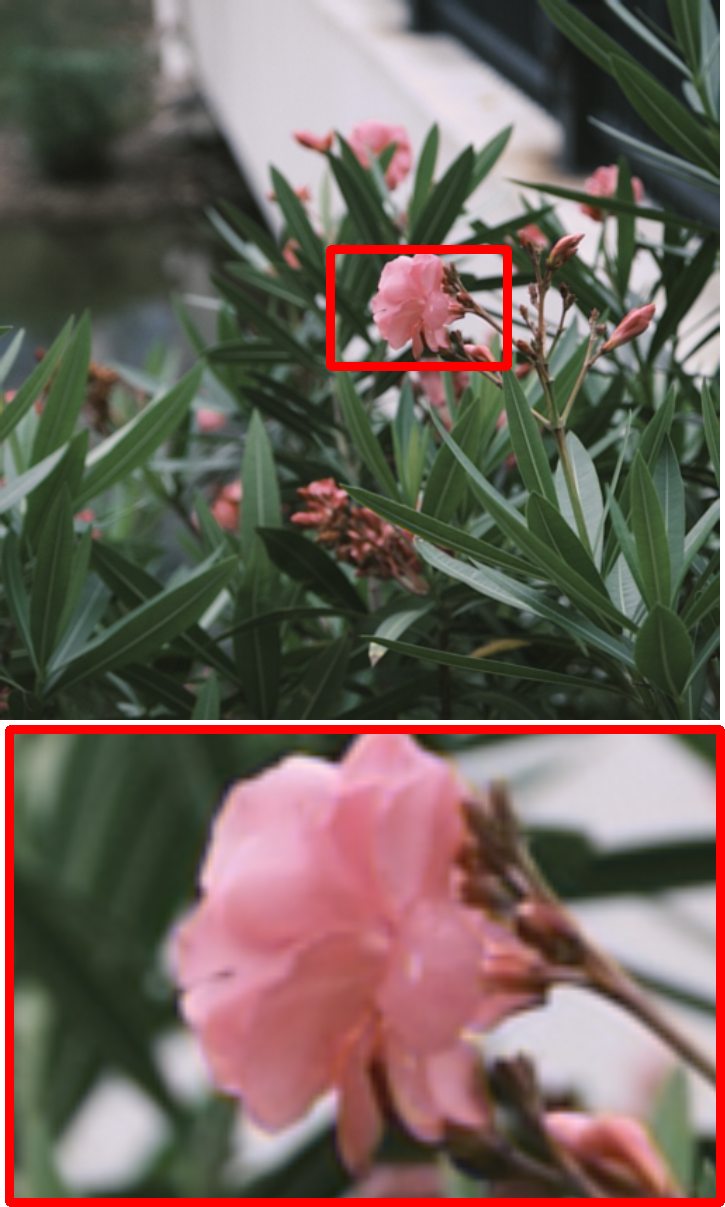}}
        \end{minipage}
        \hfill
        \begin{minipage}[b]{0.12\linewidth}
            \centering
            \centerline{\includegraphics[height=4.2cm]{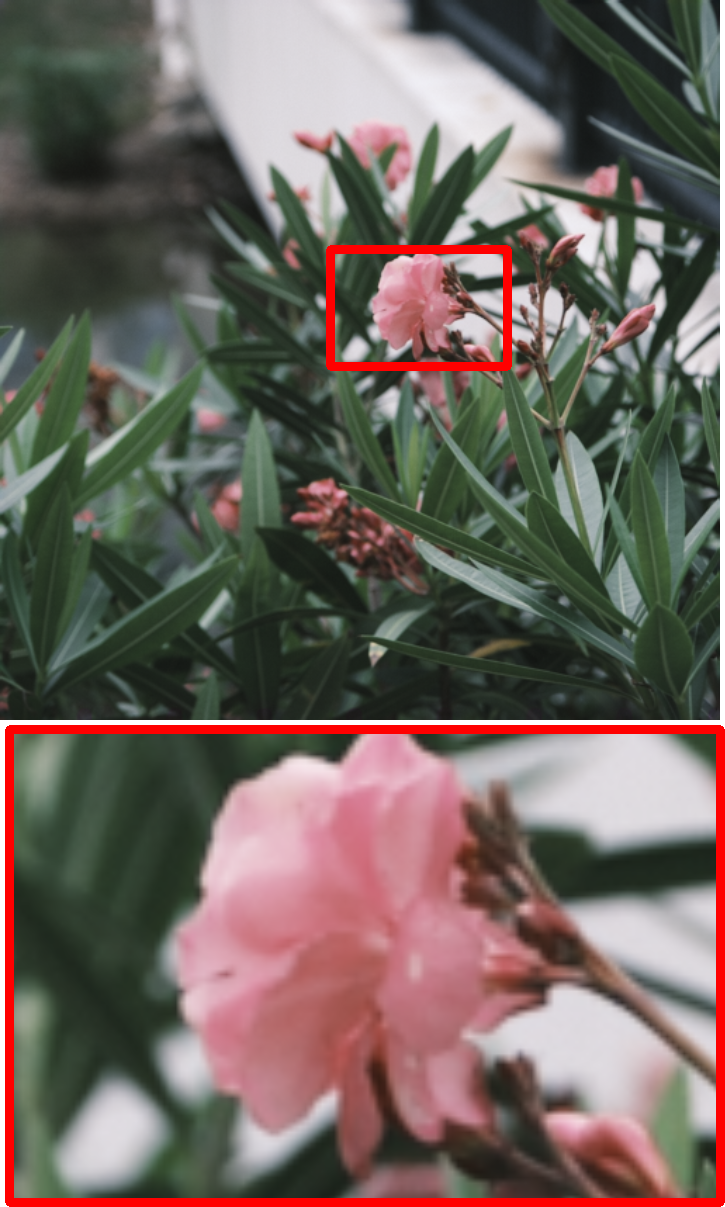}}
        \end{minipage}
        \hfill
        \begin{minipage}[b]{0.12\linewidth}
            \centering
            \centerline{\includegraphics[height=4.2cm]{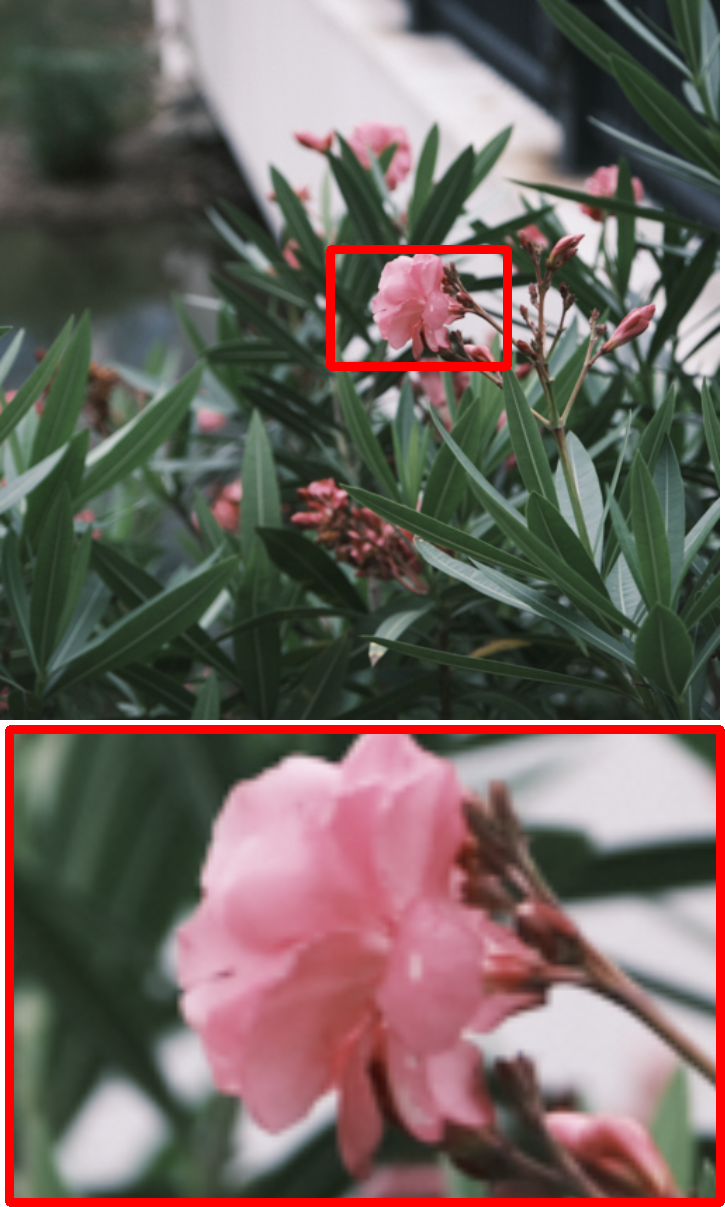}}
        \end{minipage}
    \end{minipage}
    
    \begin{minipage}[b]{1.0\linewidth}
        \begin{minipage}[b]{0.12\linewidth}
            \centering
            \centerline{\includegraphics[height=4.2cm]{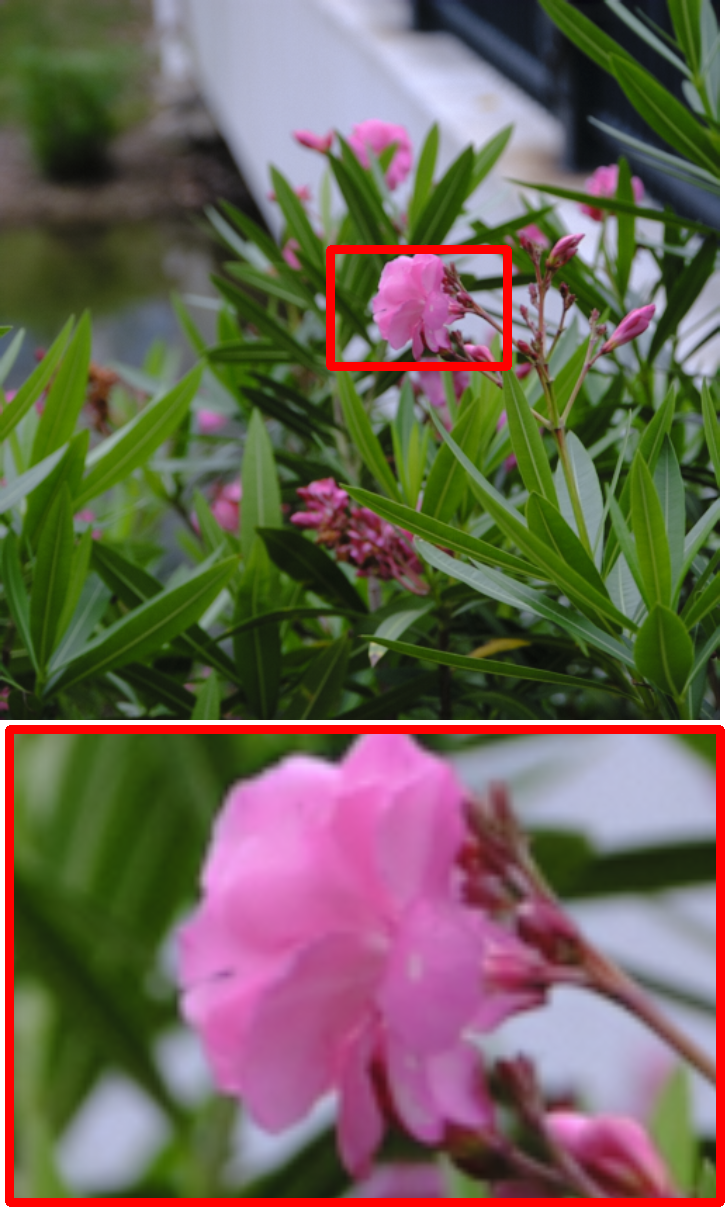}}
            \centerline{(a) Input}\medskip
        \end{minipage}
        \hfill
        \begin{minipage}[b]{0.12\linewidth}
            \centering
            \centerline{\includegraphics[height=4.2cm]{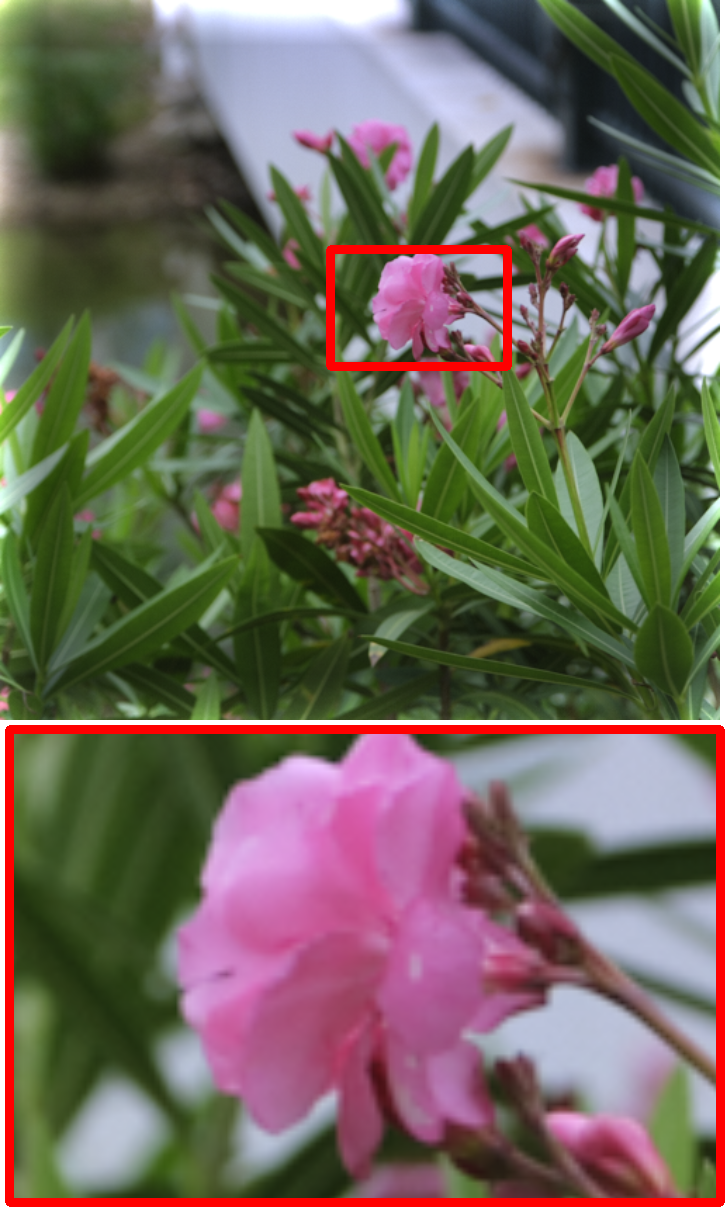}}
            \centerline{(c) UPE}\medskip
        \end{minipage}
        \hfill
        \begin{minipage}[b]{0.12\linewidth}
            \centering
            \centerline{\includegraphics[height=4.2cm]{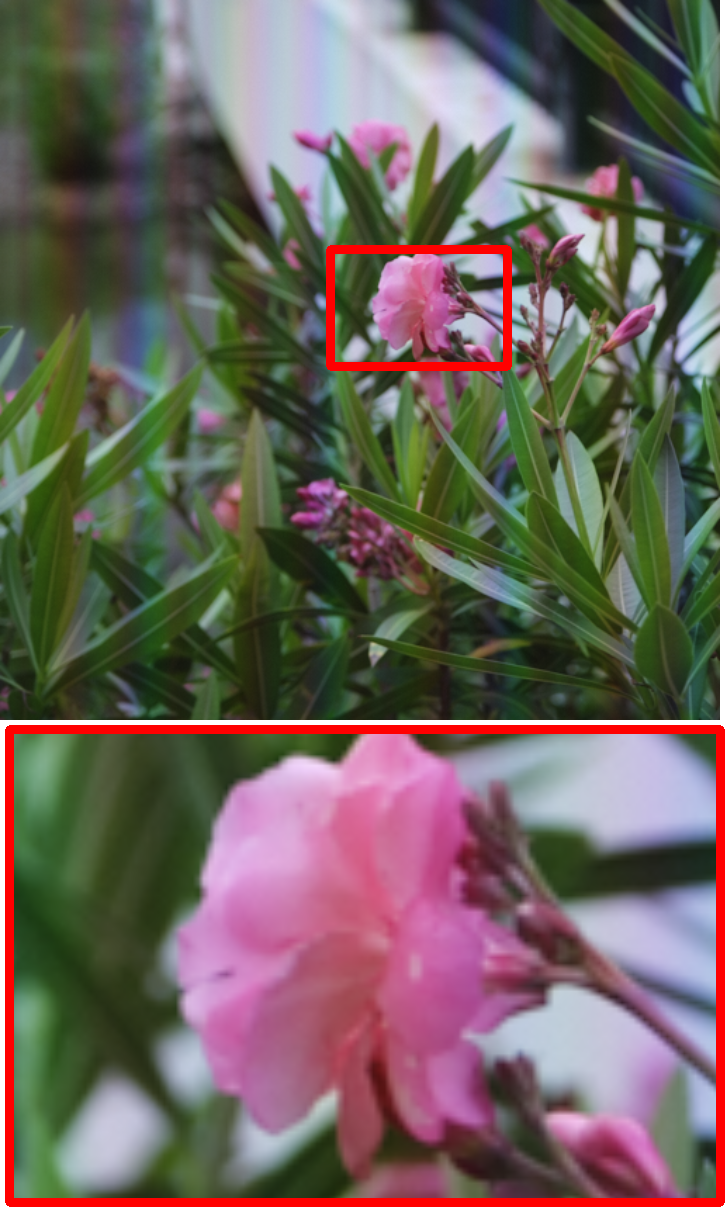}}
            \centerline{(d) STAR-DCE}\medskip
        \end{minipage}
        \hfill
        \begin{minipage}[b]{.12\linewidth}
            \centering
            \centerline{\includegraphics[height=4.2cm]{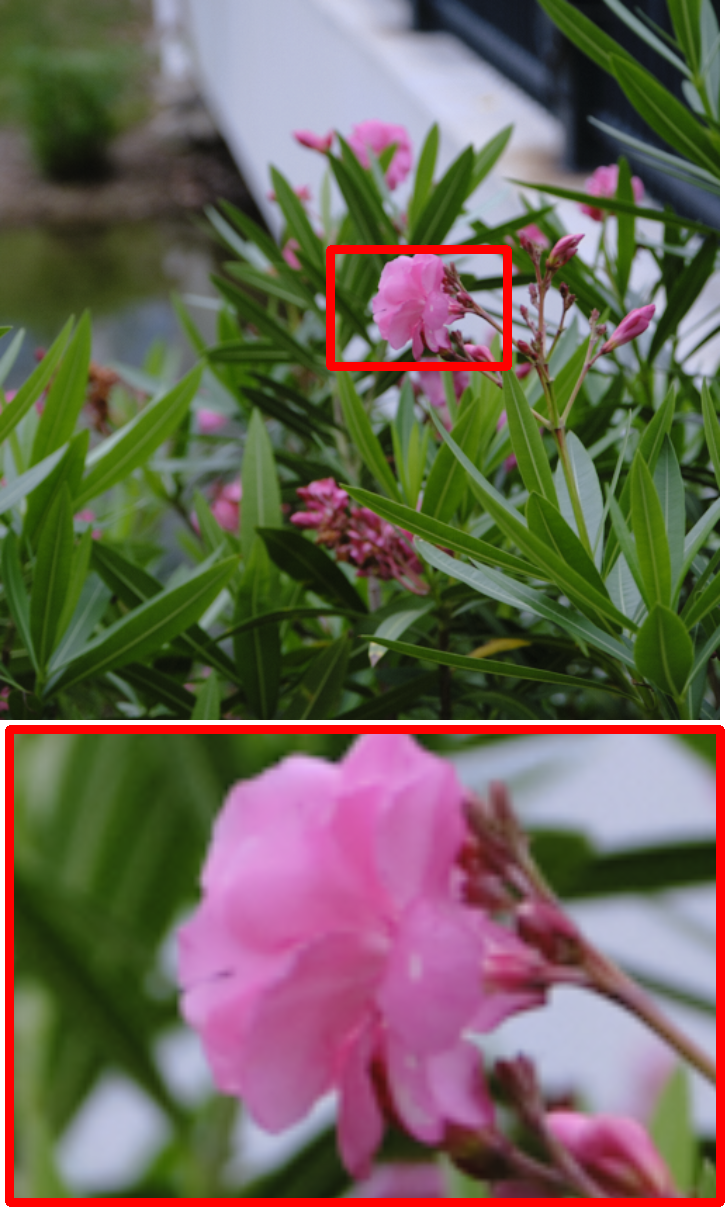}}
            \centerline{(e) 3D-LUT}\medskip
        \end{minipage}
        \hfill
        \begin{minipage}[b]{0.12\linewidth}
            \centering
            \centerline{\includegraphics[height=4.2cm]{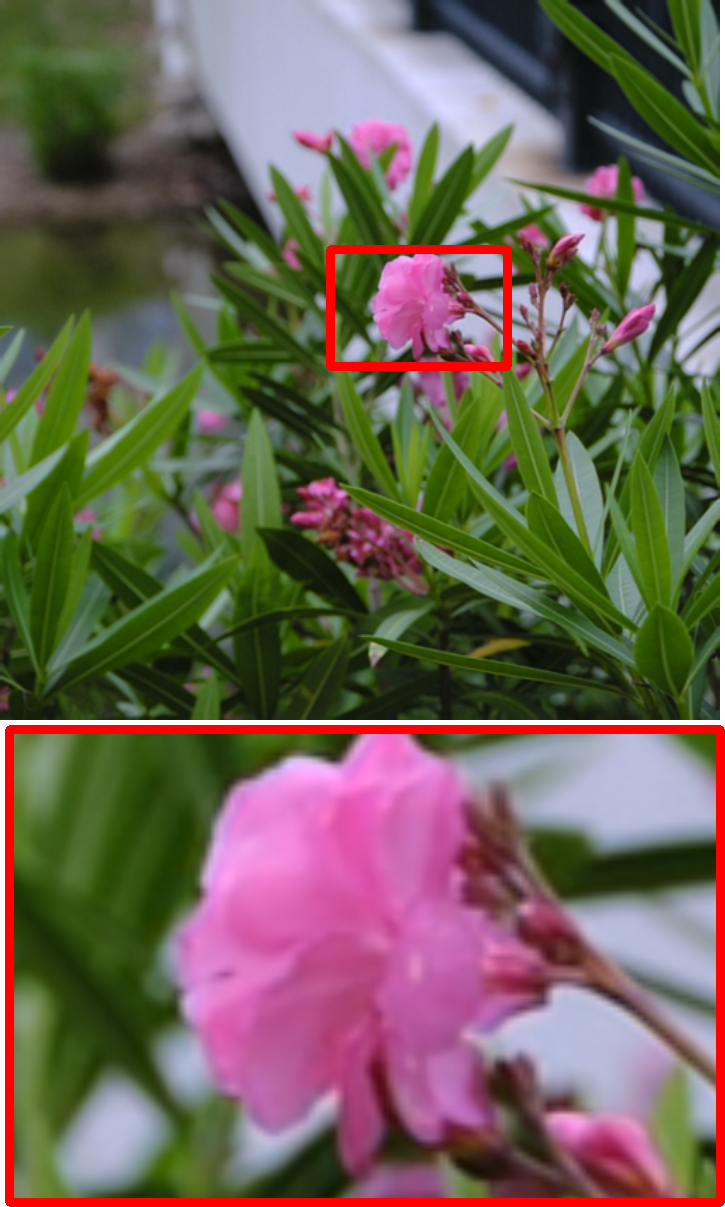}}
            \centerline{(f) LPTN}\medskip
        \end{minipage}
        \hfill
        \begin{minipage}[b]{0.12\linewidth}
            \centering
            \centerline{\includegraphics[height=4.2cm]{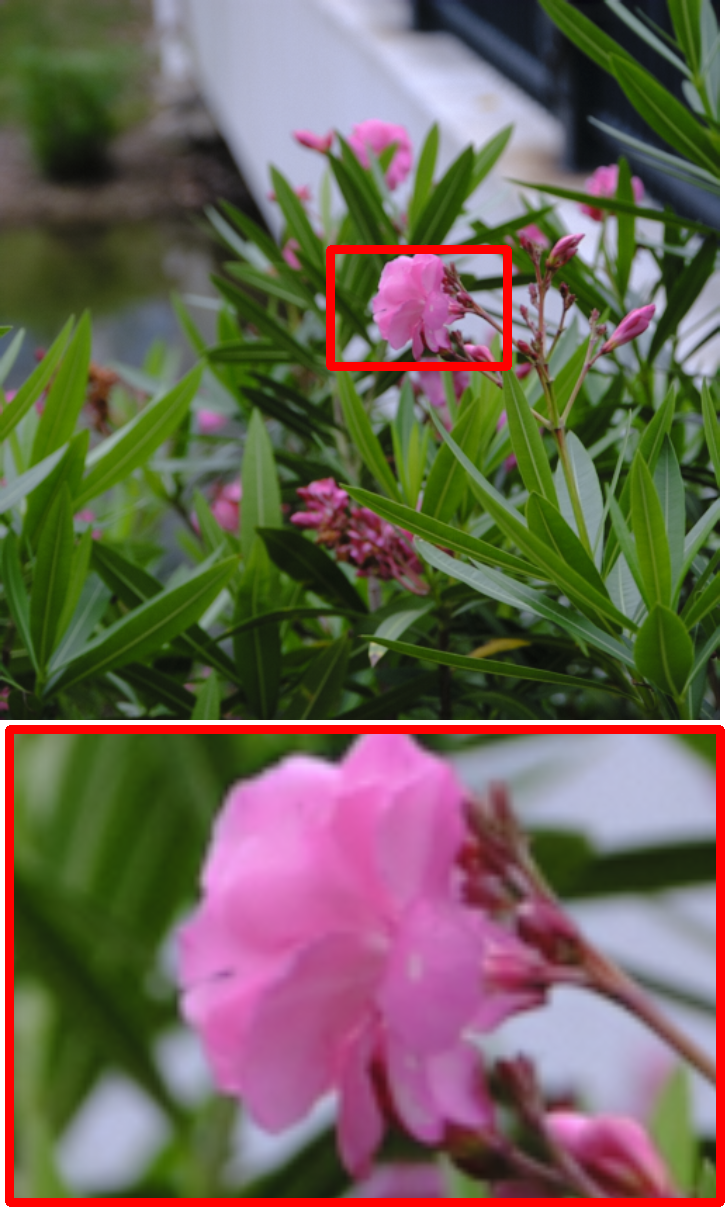}}
            \centerline{(g) Ours}\medskip
        \end{minipage}
        \hfill
        \begin{minipage}[b]{0.12\linewidth}
            \centering
            \centerline{\includegraphics[height=4.2cm]{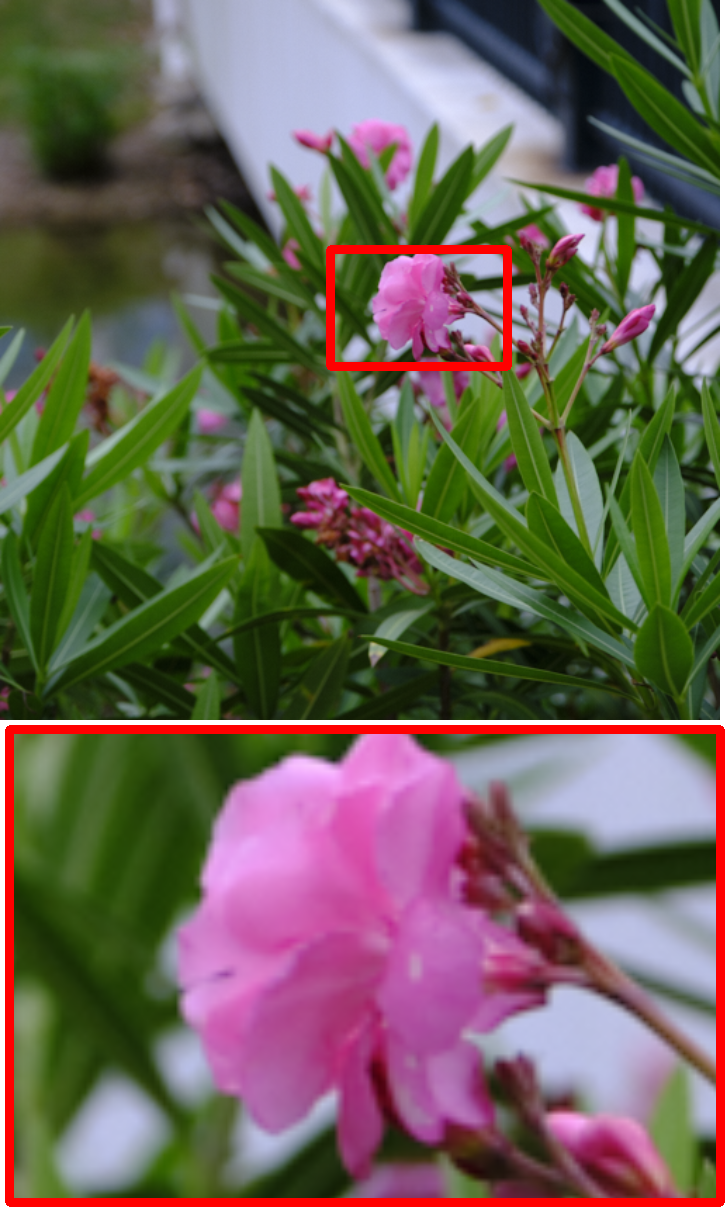}}
            \centerline{(h) Target}\medskip
        \end{minipage}
    \end{minipage}

    \caption{
    Visual comparison of different methods for image enhancement on the FilmSet dataset. Our results~(g) are visually better in color tone and details. Due to space constraints and the inaccessibility of DPE results, we reduce the number of displayed images. The input~(a) and target~(h) are the reference images from the FilmSet. Each row of images represents Cinema, ClassNeg and Velvia film style vertically.
    }
    \label{fig:film}
\end{figure*}

\begin{figure*}[!ht]
    \begin{minipage}[b]{1.0\linewidth}
        \begin{minipage}[b]{0.24\linewidth}
            \centering
            \centerline{\includegraphics[width=4.45cm]{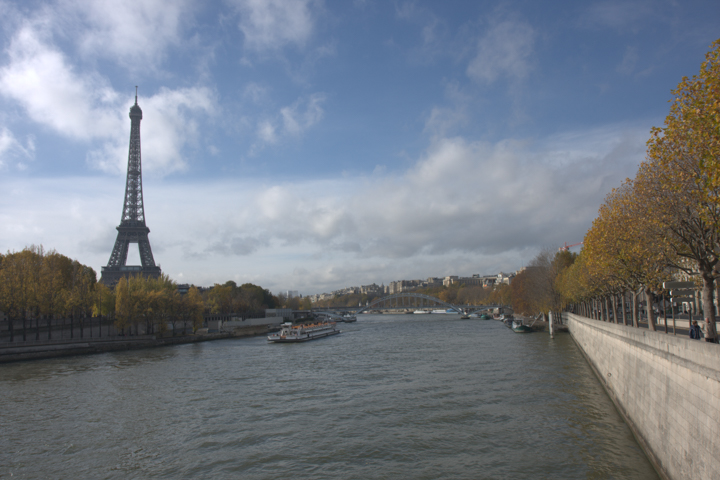}}
            \centerline{(a) Input}\medskip
        \end{minipage}
        \hfill
        \begin{minipage}[b]{0.24\linewidth}
            \centering
            \centerline{\includegraphics[width=4.45cm]{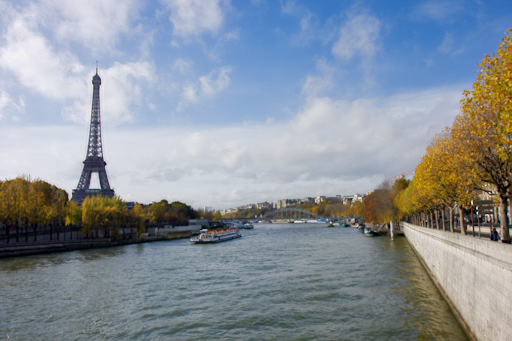}}
            \centerline{(b) DPE}\medskip
        \end{minipage}
        \hfill
        \begin{minipage}[b]{0.24\linewidth}
            \centering
            \centerline{\includegraphics[width=4.45cm]{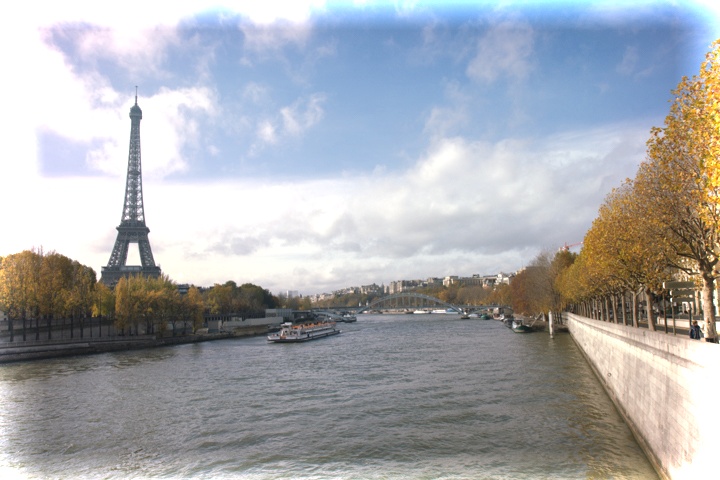}}
            \centerline{(c) UPE}\medskip
        \end{minipage}
        \hfill
        \begin{minipage}[b]{0.24\linewidth}
            \centering
            \centerline{\includegraphics[width=4.45cm]{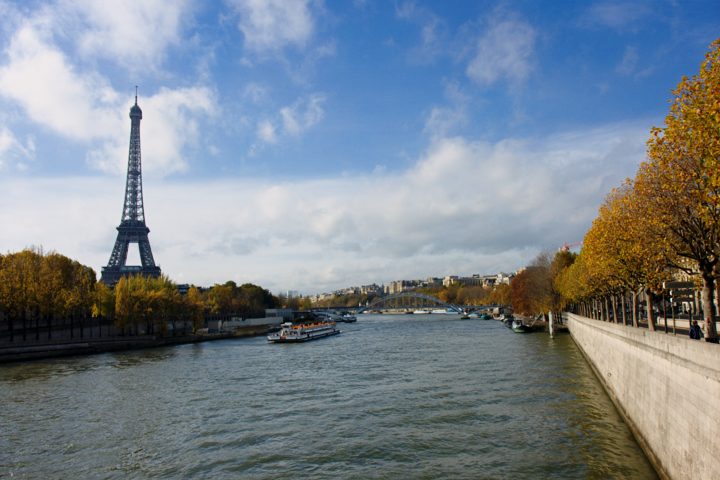}}
            \centerline{(d) LPTN}\medskip
        \end{minipage}
    \end{minipage} 

    \begin{minipage}[b]{1.0\linewidth}
        \begin{minipage}[b]{0.24\linewidth}
            \centering
            \centerline{\includegraphics[width=4.45cm]{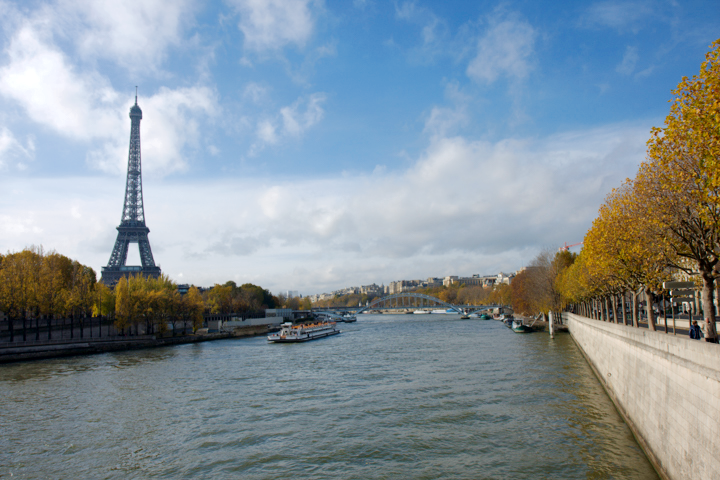}}
            \centerline{(e) 3D-LUT}\medskip
        \end{minipage}
        \hfill
        \begin{minipage}[b]{0.24\linewidth}
            \centering
            \centerline{\includegraphics[width=4.45cm]{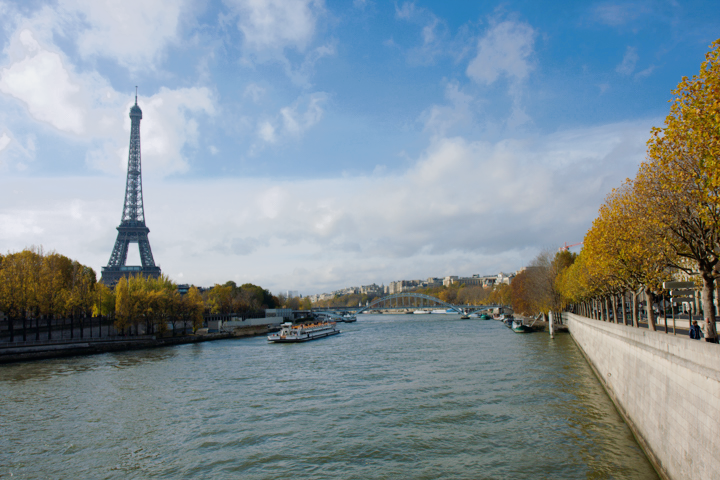}}
            \centerline{(f) SepLUT}\medskip
        \end{minipage}
        \hfill
        \begin{minipage}[b]{0.24\linewidth}
            \centering
            \centerline{\includegraphics[width=4.45cm]{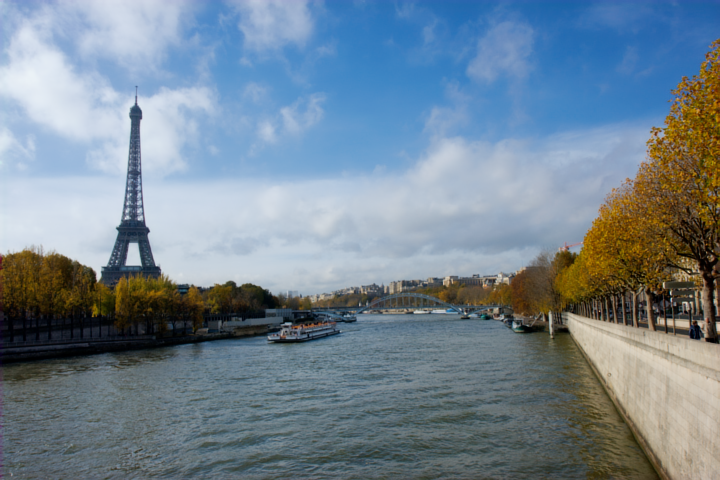}}
            \centerline{(g) Ours}\medskip
        \end{minipage}
        \hfill
        \begin{minipage}[b]{0.24\linewidth}
            \centering
            \centerline{\includegraphics[width=4.45cm]{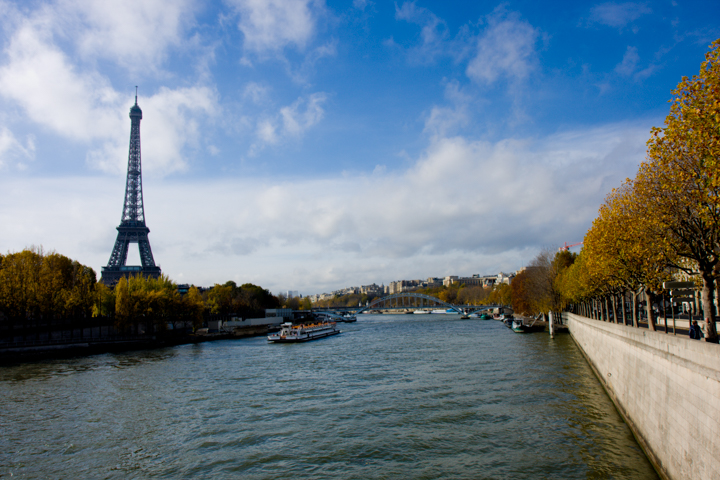}}
            \centerline{(h) Target}\medskip
        \end{minipage}
    \end{minipage} 
    
    \caption{
    Visual comparison of various image enhancement methods on the FiveK dataset. Our result~(g) is aesthetically superior in terms of color tone and specifics. The DPE~(b) and UPE~(c) results greatly deviate from the objective. Their color, exposure and reproduction of fine details are unsatisfactory. 3D-LUT~(e), SepLUT~(f) and LPTN~(d) are visually better, however the tone mapping is typically too bright or too dark compared to the target~(h).
    }
    \label{fig:fivek}
\end{figure*}
\section{Proposed Method: FilmNet}
Following the core idea of multi-frequency optimization, initially, our network splits the input picture into two distinct areas using  Laplacian Pyramid (LP): two high-frequency regions representing textures and edges, and a colored low-frequency region. The $512 \times 512$ input image is downsampled to $128 \times 128$ and sent into the Nonlinear Stylization Remapping (NSR) block to precisely adjust the color details, which substantially increases computing efficiency. The high-frequency sections are input individually into a cascade network, and a mask is learned and expanded, which saves our computation volume and enables us to optimize the high-frequency regions more effectively. Finally, these three images are recombined into a single picture $I_{out}$, downsampled to a Low-Resolution (LR) image $I_{LR}$ and fed into Triple Trilinear Regulator (TTR) alongside the $I_{out}$, and the weights of the TTR are adjusted using a CNN to produce the final film output. Each node in our network lightweightimal parameters, resulting a more efficient caltulation. The overall framework of FilmNet can be seen in Figure~\ref{fig:filmnet}.

\subsection{Multi-frequency Style Transferring}
\subsubsection{Laplacian Pyramid}
Since the characteristics of film style images are very suitable for multi-frequency enhancement and inspired by~\cite{liang2021high}, Laplacian Pyramid (LP)~\cite{burt1987laplacian} is applied in the first phase, which is a time-tested image processing method. This allows us to refine images in various frequency domains and obtain high quality results. The LP stores the picture difference between each level's blurred version and it consists of linearly decomposing a picture into a series of high- and low-frequency bands, from which the original image may be precisely rebuilt.

Given an $H \times W$ input image $I$, it creates a low-pass prediction $\hat{I} \in R^{\frac{H}{2} \times \frac{W}{2}}$ in which each pixel is a weighted average of its nearby pixels using a specified kernel. In order to provide reversible reconstruction, the LP stores the high-frequency residual $h_0$ as $h_0 = I_0 - \hat{I_0}$, where $\hat{I_0}$ represents the upsampled picture from $\hat{I}$. To further lower the sample rate and picture resolution, LP repeatedly performs the preceding procedures on $\hat{I}$ to provide a series of low-frequency and high-frequency components. The whole process can be wrote as Equation~\ref{eq:lp}:
\begin{equation}
L_{i}=G_{i}-\operatorname{PyrUp}\left(\operatorname{PyrDown}\left(G_{i}\right)\right)
\label{eq:lp}
\end{equation}
where PyrUp and PyrDown represent the upsample and downsample operations, respectively. $G_{i}$ is the source image and $L_{i}$ is the corresponding LP image.

\subsubsection{Nonlinear Stylization Remapping}
The low-frequency part $I_i$ of the input image is sent to a UNet-like architecture for detailed color transfer, namely Nonlinear Stylization Remapping (NSR), which output refined result $\hat{I_i}$. Inspired by~\cite{chen2022simple}, NSR is a simplified nonlinear network with lightweight parameters, which eliminates extraneous activation functions such as Sigmoid, Softmax, ReLU, etc. and it has been demonstrated the performance will not drop. The NSR block begins with the addition of LayerNorm inspired by~\cite{ba2016layer} to stabilize the training process, followed by two convolutions. After the deformable convolution, SimpleGate and Simplified Channel Attention (SCA) are utilized to improve the performance. 

As shown in Figure~\ref{fig:filmnet}, SimpleGate separates the features straight into two pieces along the channel dimension and multiplies them together and SCA utilizes a direct $1 \times 1$ convolution technique to transmit data across channels. SimpleGate can be described as Equation~\ref{eq:sg}:
\begin{equation}
\operatorname{SimpleGate}(\mathbf{X}, \mathbf{Y})=\mathbf{X} \odot \mathbf{Y}
\label{eq:sg}
\end{equation}
where X and Y are identically sized feature maps, $\odot$ is an element-wise multiplication.

Given a fully-connected layer $W$, $pool$ represents the global average pooling procedure that combines spatial data into channels, $*$ indicates a channel-wise multiplication, SCA can be described as Equation~\ref{eq:sca}: 
\begin{equation}
SCA(\mathbf{X})=\mathbf{X} * W \operatorname{pool}(\mathbf{X})
\label{eq:sca}
\end{equation}

\subsubsection{High-frequency Refinement}
In this section, we learn a mask on the concatenation of $[I_{i-1}, up(I_i), up(\hat{I_i})]$, as shown in Figure~\ref{fig:filmnet}, where $up(\cdot)$ indicates bilinear upsampling. This mask is gradually enlarged to improve the remaining high-frequency components based on the inherent property of LP. Given an input high-frequency image $H$ and mask $M$, the output is described as: $H_{out} = H \otimes M$, where $\otimes$ represents the pixel-wise multiplication. This is a simpler method for optimizing global correction compared to mixed-frequency images since high-frequency bands vary only little, allowing us to reduce the calculation volume~\cite{liang2021high}. 

Then, for better refinement, we introduce Multi-Scale Reconstruction Module (MSRM). There are two components in MSRM: Global-Aware Convolution and Attentive Aggregation Node~\cite{cun2020towards}. Inspired by~\cite{xu2022shadow}, Global-Aware Convolution is designed as a two-branch structure with a lightweight convolution~\cite{wu2019pay} and a standard convolution, focusing on both light and dark areas of images. The lightweight convolution seeks to learn the color mapping in the lighter area, while conventional convolution seeks to learn the color mapping in the darker part. The outputs of these two branches are then blended and added to the input characteristics using a shortcut in order to increase contextual similarity and decrease learning difficulty.

The Attentive Aggregation Node is designed for feature aggregation and attention aggregation. Each aggregation node utilizes a squeeze-and-excitation block~\cite{hu2018squeeze} to re-weight the significance of each feature channel. Then, a $3 \times 3$ convolution is used to compress the features and match the original channels. At the end of the MSRM, a spatial pooling pyramid (SPP)~\cite{he2015spatial} is introduced to facilitate the remixing of multi-context features. As shown in Figure~\ref{fig:filmnet}, the $1/r$ in SPP represents the Average Pooling with $stride = r$.

\subsection{Global Refinement}
After completing the above procedures, the three images are combined into one and sent into a lightweight TTR module. TTR has been designed to further enhance the tone of film styles. It consists of three 3-dimensional lookup table (3DLUT) weight matrices, which are used to perform tri-linear interpolation. Given a source image $I$, we first send it to an Style-Aware Adjuster. The Style-Aware Adjuster is a CNN network and the weight is borrowed from\cite{zeng2020learning}. As seen in Figure~\ref{fig:filmnet}, the Style-Aware Adjuster modifies the weight of three basis 3D LUTs on the left side based on the input LR image. Given a source image with RGB color $\left\{r_{(x, y, z)}^{I}, g_{(x, y, z)}^{I}, b_{(x, y, z)}^{I}\right\}$, a LUT is performed in order to determine its position $(x, y, z)$ in the 3D LUT lattice as Equation~\ref{eq:lut}:
\begin{equation}
x=\frac{r_{(x, y, z)}^{I}}{s}, y=\frac{g_{(x, y, z)}^{I}}{s}, z=\frac{b_{(x, y, z)}^{I}}{s}
\label{eq:lut}
\end{equation}
where $s =\frac{C_{max}}{M}$, $C_{max}$ refers to the maximum color value and $M$ indicates the number of bins in each color channel.

In the whole training phase, we use $MSE$ and $SSIM$ as loss functions as Equation~\ref{eq:mse} and~\ref{eq:ssim}:
\begin{equation}
L_{MSE}=\frac{\sum_{\mathrm{i}=1}^{n}(f(x)-y)^{2}}{n}
\label{eq:mse}
\end{equation}

\begin{equation}
L_{SSIM}=\frac{\left(2 \mu_{\mathrm{x}} \mu_{\mathrm{y}}+C_{1}\right)\left(2 \sigma_{\mathrm{xy}}+C_{2}\right)}{\left(\mu_{\mathrm{x}}^{2}+\mu_{\mathrm{y}}^{2}+C_{1}\right)\left(\sigma_{\mathrm{x}}^{2}+\sigma_{\mathrm{y}}^{2}+C_{2}\right)}
\label{eq:ssim}
\end{equation}
where $\mu_{\mathrm{x}}$ and $\mu_{\mathrm{y}}$ represent the mean of images X and Y, $\sigma_{\mathrm{x}}y$ indicates the covariance between images X and Y, $\sigma_{\mathrm{x}}$ and $\sigma_{\mathrm{y}}$ represent the standard deviation of images X and Y. Normally, $C_1=(K_1 \times L)^{2}$ and $C_2=(K_2 \times L)^{2}$ with $K_1$, $K_2$ and $L$ set to $0.01$, $0.03$ and $255$.

We set the weight of $SSIM$ function to 0.4, so our total loss function can be wrote as Equation~\ref{eq:total}:
\begin{equation}
L_{total} = L_{MSE} + 0.4 * L_{SSIM}
\label{eq:total}
\end{equation}

\begin{table*}[!ht]
\centering
\resizebox{0.7\linewidth}{!}{
\begin{tabular}{c|ccc|ccc}
\hline
                         & \multicolumn{3}{c|}{Fivek}                                                                                                     & \multicolumn{3}{c}{HDR+}                                                                                \\ \cline{2-7} 
\multirow{-2}{*}{Method} & \multicolumn{1}{c|}{{\color[HTML]{000000} PSNR$\uparrow$}} & \multicolumn{1}{c|}{SSIM$\uparrow$} & $\Delta E$$\downarrow$      & \multicolumn{1}{c|}{PSNR$\uparrow$} & \multicolumn{1}{c|}{SSIM$\uparrow$} & $\Delta E$$\downarrow$      \\ \hline
HDRNet                   & 19.93                                                      & 0.798                               & 14.42                       & 23.04                               & 0.879                               & 8.97                        \\
DPE                      & 17.66                                                      & 0.725                               & 17.71                       & 22.56                               & 0.872                               & 10.45                       \\
UPE                      & 21.88                                                      & 0.853                               & 10.80                       & 21.21                               & 0.816                               & 13.05                       \\
DeepLPF                  & 24.55                                                      & 0.846                               & 8.62                        & N/A                                 & N/A                                 & N/A                         \\
3D-LUT                   & 24.59                                                      & 0.846                               & 8.30                        & 23.54                               & 0.885                               & 7.93                        \\
STAR-DCE                 & 24.50                                                       & 0.893                               & N/A                         & 26.50                                & 0.883                               & 5.77                        \\
LPTN                   & 22.19                                                      &  0.878                               & 11.90                        & N/A                                 & N/A                                 & N/A                          \\
SepLUT                   & 25.02                                                      & 0.873                               & 7.91                        & N/A                                 & N/A                                 & N/A                         \\
Ours                     & {\color[HTML]{CB0000} \textbf{25.20}}                               & {\color[HTML]{CB0000} \textbf{0.906}}        & {\color[HTML]{CB0000} \textbf{7.62}} & {\color[HTML]{CB0000} \textbf{28.06}}        & {\color[HTML]{CB0000} \textbf{0.916}}        & {\color[HTML]{CB0000} \textbf{5.41}} \\ \hline
\end{tabular}
}
\caption{Quantitative comparisons on the MIT FiveK and HDR+ dataset of different image enhancement methods. "N/A" indicates that the result is unavailable and the top result is highlighted in red.}
\label{table:exp_1}
\end{table*}
\begin{table*}[!ht]
\centering
\resizebox{0.9\linewidth}{!}{
\begin{tabular}{c|ccc|ccc|ccc}
\hline
                         & \multicolumn{3}{c|}{Cinema}                                                                                                         & \multicolumn{3}{c|}{ClassNeg}                                                                                                       & \multicolumn{3}{c}{Velvia}                                                                                                          \\ \cline{2-10} 
\multirow{-2}{*}{Method} & \multicolumn{1}{c|}{PSNR$\uparrow$}               & \multicolumn{1}{c|}{SSIM$\uparrow$}               & $\Delta E$$\downarrow$      & \multicolumn{1}{c|}{PSNR$\uparrow$}               & \multicolumn{1}{c|}{SSIM$\uparrow$}               & $\Delta E$$\downarrow$      & \multicolumn{1}{c|}{PSNR$\uparrow$}               & \multicolumn{1}{c|}{SSIM$\uparrow$}               & $\Delta E$$\downarrow$      \\ \hline
HDRNet                   & \multicolumn{1}{c|}{35.18}                        & \multicolumn{1}{c|}{0.990}                        & 2.81                        & \multicolumn{1}{c|}{35.41}                        & \multicolumn{1}{c|}{0.988}                        & 2.19                        & \multicolumn{1}{c|}{34.37}                        & \multicolumn{1}{c|}{0.975}                        & 3.56                        \\
DPE                      & \multicolumn{1}{c|}{3.98}                         & \multicolumn{1}{c|}{0.358}                        & 47.58                       & \multicolumn{1}{c|}{3.79}                         & \multicolumn{1}{c|}{0.320}                        & 49.66                       & \multicolumn{1}{c|}{3.48}                         & \multicolumn{1}{c|}{0.313}                        & 52.12                       \\
UPE                      & \multicolumn{1}{c|}{22.81}                        & \multicolumn{1}{c|}{0.946}                        & 4.22                        & \multicolumn{1}{c|}{22.50}                        & \multicolumn{1}{c|}{0.936}                        & 4.97                        & \multicolumn{1}{c|}{22.23}                        & \multicolumn{1}{c|}{0.893}                        & 5.00                        \\
DeepLPF                  & \multicolumn{1}{c|}{36.34}                        & \multicolumn{1}{c|}{0.985}                        & 1.96                        & \multicolumn{1}{c|}{33.40}                        & \multicolumn{1}{c|}{0.978}                        & 2.43                        & \multicolumn{1}{c|}{34.06}                        & \multicolumn{1}{c|}{0.956}                        & 2.24                        \\
3D-LUT                   & \multicolumn{1}{c|}{35.49}                        & \multicolumn{1}{c|}{0.990}                        & 1.86                        & \multicolumn{1}{c|}{33.82}                        & \multicolumn{1}{c|}{0.989}                        & 1.83                        & \multicolumn{1}{c|}{34.07}                        & \multicolumn{1}{c|}{0.976}                        & 2.40                        \\
STAR-DCE                 & \multicolumn{1}{c|}{28.12}                        & \multicolumn{1}{c|}{0.949}                        & 6.91                        & \multicolumn{1}{c|}{25.54}                        & \multicolumn{1}{c|}{0.945}                        & 7.98                        & \multicolumn{1}{c|}{34.06}                        & \multicolumn{1}{c|}{0.956}                        & 2.24                        \\
LPTN                   & \multicolumn{1}{c|}{36.55}                        & \multicolumn{1}{c|}{0.985}                        & 2.12                        & \multicolumn{1}{c|}{34.22}                        & \multicolumn{1}{c|}{0.972}                        & 2.72                        & \multicolumn{1}{c|}{33.19}                        & \multicolumn{1}{c|}{0.948}                        & 3.32                        \\
SepLUT                   & \multicolumn{1}{c|}{35.82}                        & \multicolumn{1}{c|}{0.986}                        & 2.42                        & \multicolumn{1}{c|}{34.10}                        & \multicolumn{1}{c|}{0.982}                        & 2.34                        & \multicolumn{1}{c|}{32.88}                        & \multicolumn{1}{c|}{0.964}                        & 3.60                       \\
Ours                     & \multicolumn{1}{c|}{{\color[HTML]{CB0000} \textbf{40.07}}} & \multicolumn{1}{c|}{{\color[HTML]{CB0000} \textbf{0.993}}} & {\color[HTML]{CB0000} \textbf{1.61}} & \multicolumn{1}{c|}{{\color[HTML]{CB0000} \textbf{38.89}}} & \multicolumn{1}{c|}{{\color[HTML]{CB0000} \textbf{0.992}}} & {\color[HTML]{CB0000} \textbf{1.47}} & \multicolumn{1}{c|}{{\color[HTML]{CB0000} \textbf{37.60}}} & \multicolumn{1}{c|}{{\color[HTML]{CB0000} \textbf{0.981}}} & {\color[HTML]{CB0000} \textbf{2.05}} \\ \hline
\end{tabular}
}
\caption{Quantitative comparisons on the FilmSet dataset of different image enhancement methods. The top result is highlighted in red.}
\label{table:exp_2}
\end{table*}

\begin{table*}[!ht]
\centering
\resizebox{0.8\linewidth}{!}{
\begin{tabular}{c|ccc|ccc}
\hline
                         & \multicolumn{3}{c|}{Fivek}                                                                                                     & \multicolumn{3}{c}{HDR+}                                                                                \\ \cline{2-7} 
\multirow{-2}{*}{Method} & \multicolumn{1}{c|}{{\color[HTML]{000000} PSNR$\uparrow$}} & \multicolumn{1}{c|}{SSIM$\uparrow$} & $\Delta E$$\downarrow$      & \multicolumn{1}{c|}{PSNR$\uparrow$} & \multicolumn{1}{c|}{SSIM$\uparrow$} & $\Delta E$$\downarrow$      \\ 
                        \hline
D2+NSR+A                 & 25.18                                                      & 0.902                               & 7.67                        & 27.22                               & 0.905                               & 6.02                        \\
D2+NSR+TTR               & 25.14                                                      & 0.903                                & 7.63                       & 27.08                               & 0.906                               & 6.24                        \\
D2+NSR                   & 25.15                                                      & 0.901                               & 7.64                        & 26.95                               & 0.888                               & 6.65                        \\
D2+UNet+A+TTR            & 21.98                                                      & 0.856                               & 11.56                       & 21.67                               & 0.842                               & 11.51                       \\
D3+NSR+A+TTR             & 25.06                                                      & 0.899                               & 7.68                        & 26.70                               & 0.901                               & 6.78                        \\
Ours                     & {\color[HTML]{CB0000} \textbf{25.20}}                               & {\color[HTML]{CB0000} \textbf{0.906}}        & {\color[HTML]{CB0000} \textbf{7.62}} & {\color[HTML]{CB0000} \textbf{28.06}}        & {\color[HTML]{CB0000} \textbf{0.916}}        & {\color[HTML]{CB0000} \textbf{5.41}} \\ \hline
\end{tabular}
}
\caption{Ablation studies on the MIT FiveK and HDR+ dataset of different image enhancement methods. The top result is highlighted in red.}
\label{table:exp_1ab}
\end{table*}
\begin{table*}[!ht]
\centering
\resizebox{0.9\linewidth}{!}{
\begin{tabular}{c|ccc|ccc|ccc}
\hline
                         & \multicolumn{3}{c|}{Cinema}                                                                                                         & \multicolumn{3}{c|}{ClassNeg}                                                                                                       & \multicolumn{3}{c}{Velvia}                                                                                                          \\ \cline{2-10} 
\multirow{-2}{*}{Method} & \multicolumn{1}{c|}{PSNR$\uparrow$}               & \multicolumn{1}{c|}{SSIM$\uparrow$}               & $\Delta E$$\downarrow$      & \multicolumn{1}{c|}{PSNR$\uparrow$}               & \multicolumn{1}{c|}{SSIM$\uparrow$}               & $\Delta E$$\downarrow$      & \multicolumn{1}{c|}{PSNR$\uparrow$}               & \multicolumn{1}{c|}{SSIM$\uparrow$}               & $\Delta E$$\downarrow$      \\ 
\hline
D2+NSR+A                 & \multicolumn{1}{c|}{39.18}                        & \multicolumn{1}{c|}{0.992}                        & 1.61                        & \multicolumn{1}{c|}{37.40}                        & \multicolumn{1}{c|}{0.991}                        & 1.55                        & \multicolumn{1}{c|}{37.54}                        & \multicolumn{1}{c|}{0.976}                        & 2.06                        \\
D2+NSR+TTR               & \multicolumn{1}{c|}{39.46}                        & \multicolumn{1}{c|}{0.992}                        & 1.61                        & \multicolumn{1}{c|}{38.65}                        & \multicolumn{1}{c|}{0.992}                        & 1.53                        & \multicolumn{1}{c|}{37.60}                        & \multicolumn{1}{c|}{0.978}                        & 2.12                        \\
D2+NSR                   & \multicolumn{1}{c|}{{\color[HTML]{000000} 39.46}} & \multicolumn{1}{c|}{0.992}                        & 1.91                        & \multicolumn{1}{c|}{37.55}                        & \multicolumn{1}{c|}{0.990}                        & 1.71                        & \multicolumn{1}{c|}{37.49}                        & \multicolumn{1}{c|}{0.977}                        & 2.61                        \\
D2+UNet+A+TTR            & \multicolumn{1}{c|}{32.10}                        & \multicolumn{1}{c|}{0.987}                        & 3.84                        & \multicolumn{1}{c|}{30.14}                        & \multicolumn{1}{c|}{0.975}                        & 4.50                        & \multicolumn{1}{c|}{33.59}                        & \multicolumn{1}{c|}{0.969}                        & 3.52                        \\
D3+NSR+A+TTR             & \multicolumn{1}{c|}{39.29}                        & \multicolumn{1}{c|}{0.992}                        & 1.94                        & \multicolumn{1}{c|}{37.50}                        & \multicolumn{1}{c|}{0.991}                        & 1.74                        & \multicolumn{1}{c|}{37.36}                        & \multicolumn{1}{c|}{0.980}                        & 2.21                        \\
Ours                     & \multicolumn{1}{c|}{{\color[HTML]{CB0000} \textbf{40.07}}} & \multicolumn{1}{c|}{{\color[HTML]{CB0000} \textbf{0.993}}} & {\color[HTML]{CB0000} \textbf{1.61}} & \multicolumn{1}{c|}{{\color[HTML]{CB0000} \textbf{38.89}}} & \multicolumn{1}{c|}{{\color[HTML]{CB0000} \textbf{0.992}}} & {\color[HTML]{CB0000} \textbf{1.47}} & \multicolumn{1}{c|}{{\color[HTML]{CB0000} \textbf{37.60}}} & \multicolumn{1}{c|}{{\color[HTML]{CB0000} \textbf{0.981}}} & {\color[HTML]{CB0000} \textbf{2.05}} \\ \hline
\end{tabular}
}
\caption{Ablation studies on the FilmSet dataset of different image enhancement methods. The top result is highlighted in red.}
\label{table:exp_2ab}
\end{table*}

\section{Experiments}
\subsection{Experimental Setup}

\subsubsection{Datasets}
In this section, three datasets are used for training and evaluation in total: MIT-Adobe FiveK~\cite{bychkovsky2011learning}, HDR+~\cite{hasinoff2016burst} and our FilmSet. The MIT-Adobe FiveK dataset is the largest image enhancement dataset available, consisting of five retouched versions of 5,000 original pictures under varied conditions. The 3640-scene HDR+ collection from Google Camera Group for high dynamic range and low-light enhancement is a burst photography dataset. And our FilmSet is a vast high-quality dataset including over 8000 images with three distinct film genres. It is configured with 4657 training samples and 638 testing samples. For easier training and validation, all images are transformed to $512 \times 512$ resolution and standard PNG format. For FiveK and HDR+, we use the same dataset configuration as~\cite{zeng2020learning} and transform all images to the more common 480p resolution and standard PNG format.

\subsubsection{Evaluation Metrics}
In this section, we analyze frameworks utilizing the Peak Signal-to-Noise Ratio~(PSNR), Structural Similarity~(SSIM), and$\Delta E$ metrics. $\Delta E$ is a measure of color variation as experienced by humans in the CIELab color space~\cite{backhaus2011color}. Greater PNSR and SSIM values imply increased performance, whereas a lower $\Delta E$ value indicates enhanced color appearance.

\subsubsection{Implementation Details}
Our implementation is based on the PyTorch. The typical Adam optimizer with its default parameters is used to train our model by NVIDIA RTX A6000. The batch size is set to 1 and the learning rate is set to $1e-4$. Random cropping, horizontal flipping, and tweaks to brightness and saturation are used to enrich data. Visual results of FilmSet and FiveK are available in Figure~\ref{fig:film} and ~\ref{fig:fivek}.

\subsection{Comparisons with State-of-the-Arts}
A total of eight state-of-the-art methods are selected in this section: HDRNet~\cite{gharbi2017deep}, DPE~\cite{chen2018deep}, UPE~\cite{wang2019underexposed}, DeepLPF~\cite{moran2020deeplpf}, 3D-LUT~\cite{zeng2020learning}, STAR-DCE~\cite{zhang2021star}, LPTN~\cite{liang2021high} and SepLUT~\cite{yang2022seplut}. We utilized SOTA models with their provided pretrained weights in FiveK and HDR+ datasets. For FilmSet, we trained SOTA models by utilizing their own training strategies. As demonstrated in Tables~\ref{table:exp_1} and~\ref{table:exp_2}, our method exceeds others across all metrics among the three datasets.  Note that DPE produces bad results, which may be because its framework is not conducive to learning the distribution of film style, so we eliminate the visual sample of DPE. Comparing FilmSet to other datasets reveals that all approaches provide excellent results, indicating that our dataset's distribution is stable and not chaotic like the manually enhanced dataset. 

\subsection{Ablation Studies}
In this section, we separate the different parts from our architecture and set $Depth_{LP}$ to 2 and 3. We do not set $Depth_{LP}$ to 1 due to the CUDA Memory limitation. In Table~\ref{table:exp_1ab} and Table~\ref{table:exp_2ab}, ``NSR" is Nonlinear Stylization Remapping; ``D” represents the depth of LP, i.e., the $Depth_{LP}$; ``A” stands for Attentive Aggregation Node and ``TTR" is Triple Trilinear Regulator. The architecture of the best results here is $Depth_{LP}=2 + NSR + Aggregation + TTR$.

In the ablation experiment, $Depth_{LP}$ is increased to 3 and 4. The module's controlled variable experiment is conducted on D4. It is evident that the results do not improve when the $Depth_{LP}$ is raised. When the $Depth_{LP}$ is set to 4, removing the TTR module slightly diminishes the results. Likewise, deleting the A resulted in a small drop as well. When both A and TTR are eliminated, the data show a slighter decline. Following this, we substitute NSR with UNet and see a significant decline in outcomes, indicating that NSR plays a rather significant influence. In summary, increasing the $Depth_{LP}$ does not mean a better performance and every component in our architecture is helpful for improving the results.

\section{Conclusion}
In this paper, we construct a new dataset FilmSet, a large-scale and high-quality library of film styles. Our dataset consists of three distinct film types and over 5000 photos captured in the field in raw format. In order to learn the features of the FilmSet images more properly, we propose the FilmNet, a new framework based on Laplacian Pyramid for refining multi-frequency pictures and achieving high-quality results. We demonstrate that the performance of our model is superior to the state-of-the-art strategies through extensive experiments. This may facilitate the film style transferring researches in deep learning methods. 

\section*{Ethical Statement}

There are no ethical issues.

\section*{Acknowledgments}

This work was supported in part by the University of Macau under Grant MYRG2022-00190-FST, in part by the Science and Technology Development Fund, Macau SAR, under Grant 0034/2019/AMJ, Grant 0087/2020/A2 and Grant 0049/2021/A, in part by the National Natural Science Foundations of China under Grants 62172403 and in part by the Distinguished Young Scholars Fund of Guangdong under Grant 2021B1515020019.

\bibliographystyle{named}
\bibliography{ref}

\end{document}